\documentclass[preprint,authoryear,review,12pt]{elsarticle}

\usepackage{amsmath}
\usepackage{amssymb}
\usepackage{booktabs}
\usepackage{lineno}
\usepackage{algorithm}
\usepackage[noend]{algpseudocode}
\usepackage{geometry}
\usepackage{svg}
\usepackage{listings}
\usepackage{tcolorbox}
\usepackage{comment}

\usepackage{graphicx}
\usepackage{subcaption}
\usepackage{caption} 
\usepackage{lscape}
\usepackage{arydshln}

\usepackage{multirow}
\usepackage{xcolor}

\newcommand{\revision}[1]{#1}

\newcommand{\secondrevision}[1]{#1}

\definecolor{roomcolor}{HTML}{bfdae5}     % light blue
\definecolor{staircolor}{HTML}{d6a33b}    % golden yellow
\definecolor{doorcolor}{HTML}{765241}     % brown

\DeclareRobustCommand{\legendcircle}[1]{%
  \tikz[baseline=-0.7ex]{
    \node[circle, draw=black, fill=#1, inner sep=0pt, minimum size=1.5ex] (c) {};
  }%
}

\usepackage{pgfplots}

\definecolor{erdline}{HTML}{8C8C8C}
\definecolor{erdtype}{HTML}{6B6B6B}
\definecolor{erdkey}{HTML}{222222}

\usepackage{setspace}

\usepackage{url}

\usepackage{enumitem}
\definecolor{codebg}{RGB}{245,247,250}      % light background
\definecolor{framegray}{RGB}{180,185,190}   % subtle frame
\definecolor{textgray}{RGB}{30,30,30}       % dark text
\definecolor{keywordcolor}{RGB}{0,102,204} % for SQL/CYPHER keywords
\definecolor{commentcolor}{RGB}{128,128,128} % for analysis comments
\definecolor{stringcolor}{RGB}{153,0,0}    % for strings/numbers

\lstdefinestyle{querylog}{
    basicstyle=\ttfamily\footnotesize\color{textgray},
    keywordstyle=\color{keywordcolor}\bfseries,
    commentstyle=\color{commentcolor}\itshape,
    stringstyle=\color{stringcolor},
    breaklines=true,
    frame=single,
    backgroundcolor=\color{codebg},
    rulecolor=\color{framegray},
    framesep=12pt,
    xleftmargin=15pt,
    xrightmargin=15pt,
    framexleftmargin=0pt,
    framexrightmargin=0pt,
    aboveskip=15pt,
    belowskip=15pt,
    columns=fullflexible,
    keepspaces=true,
    showstringspaces=false,
    rulecolor=\color{framegray},
    captionpos=b
}

\journal{Elsevier}
\pgfplotsset{compat=1.18}

\begin{document}
\begin{frontmatter}



\title{IfcLLM: Natural Language Querying of IFC Models through Complementary Relational and Graph Representations}


\author[grid]{Rabindra Lamsal\corref{cor1}}
\ead{r.lamsal@unsw.edu.au}

\author[grid]{Sisi Zlatanova}
\ead{s.zlatanova@unsw.edu.au}

\author[grid]{Haowen Xu}
\ead{haowen.xu1@unsw.edu.au}

\author[civil]{Yafei Sun}
\ead{yafei.sun@unsw.edu.au}

\author[civil]{Xuesong Shen}
\ead{x.shen@unsw.edu.au}

\affiliation[grid]{organization={GRID Lab, School of Built Environment, The University of New South Wales},
            city={Sydney},
            country={Australia}}

\affiliation[civil]{organization={School of Civil and Environmental Engineering, The University of New South Wales},
            city={Sydney},
            country={Australia}}

\cortext[cor1]{Corresponding author}
%% Abstract
\begin{abstract}
The Industry Foundation Classes (IFC) standard is central to building data exchange across the lifecycle, from design and construction to facility management and Digital Twin integration. In operational settings, stakeholders increasingly require access to building information without specialist knowledge of IFC's complex, deeply nested schema, motivating natural language interfaces. Existing Large Language Model (LLM)-based querying approaches typically rely on a single data representation, which is not equally suited to attribute retrieval and spatial reasoning. We present IfcLLM, a framework that combines complementary relational and graph representations, routing each query type to the more suitable backend. An LLM agent integrates both through iterative retry-and-refine reasoning, recovering from failures without user input. Evaluated on 30 query scenarios across three IFC models, the top-performing configuration achieved 96\% accuracy. Results with open-weight LLMs further demonstrate the potential for local deployment in data-sensitive AEC settings.\end{abstract}



%% Keywords

\end{frontmatter}


\section{Introduction}
\label{sec:introduction}

The Industry Foundation Classes (IFC) standard provides a software-neutral, open schema for representing the geometry, semantics, and relationships of building elements throughout the entire life cycle of a built environment \citep{eastman2011bim,wang2019integration,ifc}. Beyond design and construction workflows, IFC is increasingly being used during the operational phase for applications such as facility management \citep{marmo2020building}, indoor navigation \citep{isikdag2013bim,liu2021indoor}, and simulation and integration with Digital Twin platforms \citep{aleksandrov2024bim,xu2026towards}. As IFC becomes a central medium for data exchange in these settings, its user base is expanding well beyond architects and structural engineers. Facility managers, urban analysts, emergency planners, and IoT system integrators increasingly rely on IFC models to access and analyze building information \citep{eastman2011bim}, yet most of these users lack the expertise to navigate IFC's complex and deeply nested schema or to formulate structured queries. This issue creates a significant barrier to the effective use of IFC data.

These diverse stakeholders require fundamentally different types of information. A facility manager may need the thermal transmittance of external walls; an emergency planner may seek the shortest navigable route between two spaces. These examples illustrate two distinct query types: attribute retrieval and spatial traversal. Formulating either type of query against the IFC schema requires specialist knowledge that most operational users do not have, and the barrier is further intensified by the large and structurally heterogeneous nature of real-world IFC models \citep{shi2018ifcdiff}. This motivates the development of intuitive natural language querying interfaces that hide the complexity of the underlying IFC schema from users.

Large language models (LLMs) \citep{zhao2023survey,achiam2023gpt} have emerged as a powerful mechanism for translating natural language requests into executable queries over structured data. However, applying LLMs directly to IFC data faces two fundamental challenges that must be resolved before a reliable querying interface can be built. The first challenge is \textit{size}. Real-world IFC files encode rich building representations including spatial hierarchy, physical elements, material properties, and geometric data \citep{ifc}. Even relatively small IFC models may expand to very large token counts\footnote{For instance, the raw text of a simple IFC model like \textit{FZK-Haus} (shown in Figure \ref{fig:hauz-ifc}) tokenizes to $\approx$1.7 million tokens with the GPT-4o tokenizer.}, often exceeding the practical context window of contemporary LLMs. Although recent models support increasingly long contexts, processing entire IFC files for every query remains computationally expensive and unsuitable for interactive use \citep{liu2024lost}. A compact, structured representation of the IFC file is therefore preferable. The second challenge is the \textit{schema}. The IFC schema is large and specialized, with deeply nested entities, recursive placement hierarchies, and domain-specific terminology \citep{eastman2011bim,ifc}. Although LLMs are trained on large and diverse text corpora \citep{zhao2023survey}, the IFC structure is not naturally aligned with conversational reasoning. These two challenges together motivated exploring mappings into structured representations that LLMs can query through standard, well-established query languages.

Transforming IFC data into queryable representations is a well-established practice. Relational mappings of the IFC schema have existed since the early 2000s \citep{adachi2001ifcmodelserver,you2004relational}, and graph-based representations have been proposed to support relationship-based reasoning \citep{zhu2023ifc}. More recently, these representations have been combined with LLMs under a retrieval-augmented generation (RAG) paradigm \citep{lewis2020retrieval, gao2023retrieval}, demonstrating the feasibility of natural language IFC querying in practice \citep{iranmanesh2025llm,pacheco2024bimconverse,pan2026llm}. \revision{This feasibility arises from RAG reframing the problem: instead of ingesting the raw IFC file, the LLM receives compact schema summaries and generates SQL or Cypher queries over the structured backend, grounding its responses in the retrieved results. This approach directly addresses the size and schema complexity challenges identified above.} However, existing systems rely on a single data representation, most commonly a graph. Attribute-centric queries such as retrieving element properties, counting elements per floor, or filtering elements by material are more naturally expressed through relational databases and SQL. Moreover SQL and relational databases are extensively represented in LLM training data, supporting more reliable translation from natural language to queries \citep{hong2025next,liu2025survey}. In contrast, spatial queries such as identifying connected spaces or computing shortest paths, are more concisely expressed through graph query languages \citep{francis2018cypher,shang2025survey} than through their recursive SQL equivalents. Combining both representations therefore allows each query class to be processed by the more suitable backend, reducing query complexity and lowering token consumption.

A further consideration concerns the choice of LLM used to orchestrate such representations. LLMs are typically accessed either as proprietary, API-based services (e.g., GPTs, Claude, Gemini) or as open-weight models that can be run on local infrastructure (e.g., Llama, Qwen, GPT-OSS). Most existing LLM-based IFC querying systems \citep{pacheco2024bimconverse,iranmanesh2025llm,Avgoren2025enhancing} adopt the former, transmitting prompts, schema summaries, and query results to external endpoints. This is significant for the AEC sector, where IFC models often encode commercially or operationally confidential information, and organisations may be unwilling or unable to share such data externally. Open-weight LLMs avoid this by allowing both the model and the structured representations it queries to remain within an organisation's own network, and recent releases such as GPT-OSS have substantially narrowed the capability gap with proprietary models for structured reasoning and query generation \citep{agarwal2025gpt}. This motivates a design requirement: an open-weight LLM should be capable of orchestrating both representations described above, enabling a fully locally deployable querying pipeline that preserves data privacy.

Building on these considerations, we present IfcLLM, a framework that combines a relational representation for attribute-centric queries with a graph representation for spatial reasoning. An LLM agent integrates both through iterative retry-and-refine reasoning, enabling natural language querying of IFC models without requiring users to understand the underlying schema or formal query languages. We implement IfcLLM as a reproducible end-to-end system using an open-weight LLM, covering IFC transformation, query generation, retrieval, and response synthesis, supporting local deployment as motivated earlier. We evaluate the framework on IFC models of varying size using 30 query scenarios covering common operational building management tasks, providing initial empirical evidence of the potential benefits of this hybrid representation for LLM-based IFC querying within the evaluated scope. \revision{We position these contributions relative to the reviewed literature and state the research questions addressed in this work later in Section \ref{sec:research-gap}.}

The remainder of the paper is organized as follows. Section \ref{sec:related-work} reviews the relevant literature; Section \ref{sec:framework} introduces the proposed framework; Section \ref{sec:evaluation-setup} discusses the implementation and the evaluation setup; Section \ref{sec:tests-evaluation} presents the test and evaluation results; Section \ref{sec:discussion} provides discussion and future work; and Section \ref{sec:conclusion} concludes the paper.

\section{Related Work}
\label{sec:related-work}
Efforts to make IFC data more accessible have progressed along several directions. This section reviews the relevant literature in three parts. First, we examine approaches for transforming IFC data into structured, queryable representations. Second, we discuss how these representations are accessed through natural language interfaces. Third, we identify the research gaps that motivate the present work.

\subsection{From IFC Parsing to Queryable Representations}
\label{sec:ifc-rep}
Working with IFC data programmatically requires parsing the native IFC representation, which encodes a hierarchically nested schema of entities and relationships. Several open-source libraries address this. IfcOpenShell \citep{ifcopenshell} is the most widely adopted toolkit, supporting both semantic entity extraction and geometry processing. IFC++ \citep{ifcplusplus} targets performance-critical C++ environments, while the xbim toolkit \citep{xbim} is tailored for .NET workflows. Lightweight web-oriented parsers include Web-IFC \citep{webifc} and the Ara 3D IFC Toolkit \citep{ara3d}. Despite their availability, most parsers do not natively transform IFC into structured representations required for scalable querying or LLM-based reasoning.

To support scalable querying, a substantial body of work has explored mapping IFC data into database-backed representations. Early approaches focused on translating the IFC EXPRESS schema into relational database schemas \citep{adachi2001ifcmodelserver,you2004relational}, enabling storage and querying through SQL. Subsequent work has proposed more application-specific relational transformations. For example, \citep{isikdag2013bim} introduced a BIM-derived relational schema tailored for indoor navigation, while \citep{solihin2017simplified} proposed a star-schema representation to support spatial queries. Other domain-specific systems include BIM-based cost estimation \citep{alzraiee2022cost} and facility management frameworks integrating BIM with big data analytics and NoSQL databases \citep{demirdougen2023bim}. Object-oriented database approaches \citep{li2016object}, such as BIMserver \citep{beetz2010bimserver}, have also been explored.

Beyond building-centric representations, studies have investigated integrating IFC with other domain-specific schemas to support broader analysis \citep{hijazi2010initial,stouffs2018achieving,biljecki2021extending}. In the geospatial domain, IFC is often mapped to standards such as CityGML \citep{groger2012citygml,diakite2022ifc2indoorgml} via Application Domain Extensions (ADEs), typically implemented on spatial database systems such as PostGIS or Oracle Spatial. Similarly, IFC spaces have been mapped into cadastral and land administration schemas \citep{alattas2021mapping}.

Graph-based representations have gained increasing attention due to their ability to explicitly model relationships between building elements. Knowledge graphs represent entities (e.g., IfcWall, IfcDoor) and their relationships (e.g., IfcRelSpaceBoundary, IfcRelContainedInSpatialStructure), enabling semantic querying and relationship-based reasoning. IFC-related graphs fall into two categories: (i) graphs that encode the full semantic hierarchy of the IFC model, and (ii) graphs that target a specific use case. IFC-GRAPH \citep{zhu2023ifc} is a notable pipeline for transforming the complete IFC schema into property graphs, making IFC data easier to query \citep{zhu2024cypher4bim}. Recent work has explored simplified graph representations instead of full-schema graphs. For instance, \citep{zhu2025semantics} proposed a semantic-based connectivity graph for indoor pathfinding, constructed from predefined IFC relationships to identify boundaries between spaces. Also, \citet{Avgoren2025enhancing} proposed reducing IFC relationships to a smaller set of core types (e.g., \textit{hasStorey}, \textit{hasSpace}, \textit{containsElements}, \textit{hasBoundary}, \textit{adjacentSpace}) to improve usability and reduce reasoning complexity for LLMs. \revision{IFC has also been represented using semantic web standards such as the ifcOWL ontology \citep{pauwels2016express} and Linked Building Data (LBD) vocabularies \citep{rasmussen2020bot}. These ontologies and vocabularies can be used to represent IFC data as RDF graphs, providing a semantic-web alternative to property-graph representations.} Taken together, this body of work shows that transforming IFC data into either relational or graph-based representations is now a mature and well-supported practice.

\subsection{Natural Language Interfaces}

Early work on natural language interfaces (NLIs) for building models focused on mapping natural language queries onto structured-query backends without LLMs. \citep{lin2016natural} introduced a system that translates queries into MongoDB operations (a document-oriented, noSQL backend) using the buildingSMART IFD Dictionary \citep{bimDict}; however, this approach provides limited support for basic attribute constraints such as numeric comparisons. BIMASR \citep{shin2021bimasr} supports voice-based commands over a relational BIM database but is limited to basic element retrieval and property modification. Similar dialogue-based systems have been introduced \citep{wang2021framework, wang2022nlp} that operate over IFC file extracts and focus primarily on attribute extraction rather than spatial reasoning.

Subsequent approaches incorporated semantic web technologies, shifting the target backend towards graph-based representations. Ontology-driven NLIs \citep{yin2023ontology} translated natural language queries into SPARQL (a query language for RDF graphs) through semantic parsing, enabling multi-constraint querying over a graph-structured ontology; however, these systems often struggle with ambiguity, complex grammatical structures, and spatial reasoning. More recently, \citep{liu2025bimcoder} proposed BIMCoder, which translates natural language queries into structured query statements (BIMserver compatible) using pre-trained language models, reinforcing the importance of backend-specific structured representations for querying building models.

A more directly related line of work applies retrieval-augmented generation (RAG) techniques \citep{lewis2020retrieval, gao2023retrieval} to querying building models, and does so almost exclusively over graph representations. \cite{iranmanesh2025llm} transform IFC models into a Neo4j property graph and guide GPT-4o to generate Cypher queries by providing the graph schema, formatting constraints, and example queries in the prompt. BIMConverse \citep{pacheco2024bimconverse} follows a similar strategy over a labelled property graph, supplying an LLM with schema details, example queries, and task-specific guidance alongside the user query. Similarly, \cite{Avgoren2025enhancing} constructs a simplified knowledge graph (Section \ref{sec:ifc-rep}), while \cite{pan2026llm} propose a multi-agent framework in which LLM agents query a graph-based Digital Twin. The choice of LLM in these systems has direct implications for the data sensitivity concerns raised in Section \ref{sec:introduction}. These existing systems appear to rely on proprietary, cloud-hosted LLMs, requiring the graph schema, prompts, and intermediate results to be transmitted externally. This is a more significant concern for IFC data than for the general text corpora typically used to evaluate LLMs: unlike openly available web text, real-world IFC models are rarely released publicly, since they encode confidential building designs, and the limited set of openly available reference models (e.g., FZK Haus and the buildingSMART sample models used in this paper) are themselves simplified cases rather than representative of real commercial projects. This scarcity of shareable data, combined with the confidentiality of real deployments, reinforces the case for LLM-based IFC querying systems that can operate entirely with an open-weight, locally hosted LLM, as adopted in this work.

\subsection{Research Gap}
\label{sec:research-gap}
The reviewed literature reveals the following observations.

First, existing IFC-to-database transformations tend toward two extremes. At one end, schema-equivalent tools such as Ifc2Sql \citep{ifcopenshell} preserve the full structure of the IFC schema, including its nested entity hierarchies and hundreds of entity-specific tables. Querying this structure requires extensive multi-hop joins through IFC-specific relationship and property-set tables, along with a large schema summary, increasing both token cost and the likelihood of incorrect query generation. Consequently, this approach is not well suited to LLM-based querying. At the other end, application-specific relational schemas (e.g., \cite{isikdag2013bim,solihin2017simplified}) and LLM-oriented graph representations (e.g., \cite{Avgoren2025enhancing,iranmanesh2025llm}) adopt more compact schemas, but each targets a single representation tailored to its specific implementation. Full-schema graph representations such as IFC-GRAPH capture the complete model but incur substantial preprocessing costs \citep{zhu2023ifc}. To our knowledge, no existing pipeline automatically generates complementary, selectively scoped relational and graph representations from a single IFC source in a form suitable for LLM-based querying, while preserving the semantic, hierarchical, and geometric information needed for attribute-based and spatial reasoning.

Second, recent LLM-based IFC querying systems converge on a single representation, which is mainly graph-based (refer to Table \ref{tab:comparison}): majority of them query a property graph using Cypher \citep{francis2018cypher}. As noted in Section \ref{sec:introduction}, while such graphs are well suited to spatial reasoning, attribute-centric queries are more naturally expressed through relational databases and benefit from SQL's stronger representation in LLM training data \citep{hong2025next,liu2025survey}. Conversely, relational-only systems can answer spatial queries but, as we show in Section \ref{ablation}, require recursive SQL formulations that are substantially more complex and token-costly than equivalent graph traversals. This complementarity has not been exploited within a single LLM-orchestrated framework: to our knowledge, no existing system combines both representations and routes each query to its more suitable backend for token-efficient query generation. \revision{A third observation is that existing systems adopt either one-shot query generation \citep{massafra2024building,pacheco2024bimconverse,iranmanesh2025llm,pan2026llm} or iterative reasoning \citep{Avgoren2025enhancing}, without isolating the effect of reasoning strategy from the effect of the underlying data representation. Relatedly, while local deployment motivates the use of open-weight LLMs in this work, no existing study has examined how the choice of open-weight LLM affects querying accuracy, efficiency, or failure behaviour within such a framework.}

\begin{table}
\footnotesize
\centering
\caption{\revision{Comparison of IfcLLM with related LLM-based approaches for natural language IFC/BIM querying. ``One-shot'' denotes single-pass query generation; ``iterative'' denotes retry-and-refine reasoning.}}
\label{tab:comparison}
\begin{tabular}{p{4cm} p{2.3cm} p{3cm} p{1.9cm} p{2.1cm}}
\toprule
\textbf{Approach} & \textbf{Representation} & \textbf{Query language} & \textbf{Reasoning} & \textbf{LLM type}\\
\midrule
\citep{massafra2024building} & Graph & JSON Object & One-shot & Proprietary \\
\citep{pacheco2024bimconverse} & Graph & Cypher & One-shot & Proprietary\\
\citet{iranmanesh2025llm} & Graph & Cypher & One-shot & Proprietary\\
\citet{Avgoren2025enhancing} & Graph & Cypher & Iterative & Proprietary\\
\citet{pan2026llm} & Graph (Digital Twin) & Cypher & One-shot & Proprietary\\
\midrule
\textbf{IfcLLM (ours)} & \textbf{Hybrid (relational\,+\,graph)} & \textbf{SQL\,+\,Cypher} & \textbf{Iterative} & \textbf{Open-weight (+ proprietary baseline)} \\
\bottomrule
\end{tabular}
\end{table}

\revision{These observations motivate the following research questions, which this work addresses through the design, implementation, and evaluation of IfcLLM:}

\revision{\textbf{RQ1}: To what extent does combining relational and graph representations improve the effectiveness and efficiency of natural-language querying of IFC models compared with single-representation alternatives?}

\revision{\textbf{RQ2}: Does the iterative retry-and-refine reasoning affect the correctness of LLM-generated structured queries, and the ability to recover from failed queries, compared with one-shot reasoning?}

\revision{\textbf{RQ3}: Does the choice of open-weight LLM affect the accuracy, efficiency, and failure characteristics of such an IFC querying system?}

These questions motivate the design of IfcLLM, described in the following section.

\section{Proposed Framework}
\label{sec:framework}

This section presents \textit{IfcLLM} (\textbf{I}ndustry \textbf{F}oundation \textbf{C}lasses \textbf{L}arge \textbf{L}anguage \textbf{M}odel), a framework for natural language interaction with IFC-based BIM models. The framework enables an LLM to query and reason over building data without requiring direct interaction with the full native IFC schema. Instead of treating an IFC file as a monolithic data source, IfcLLM transforms building information into complementary relational and graph representations, each optimised for different types of queries and reasoning tasks.

\subsection{Architecture} 

\begin{figure}
    \centering
    \includegraphics[width=\textwidth]{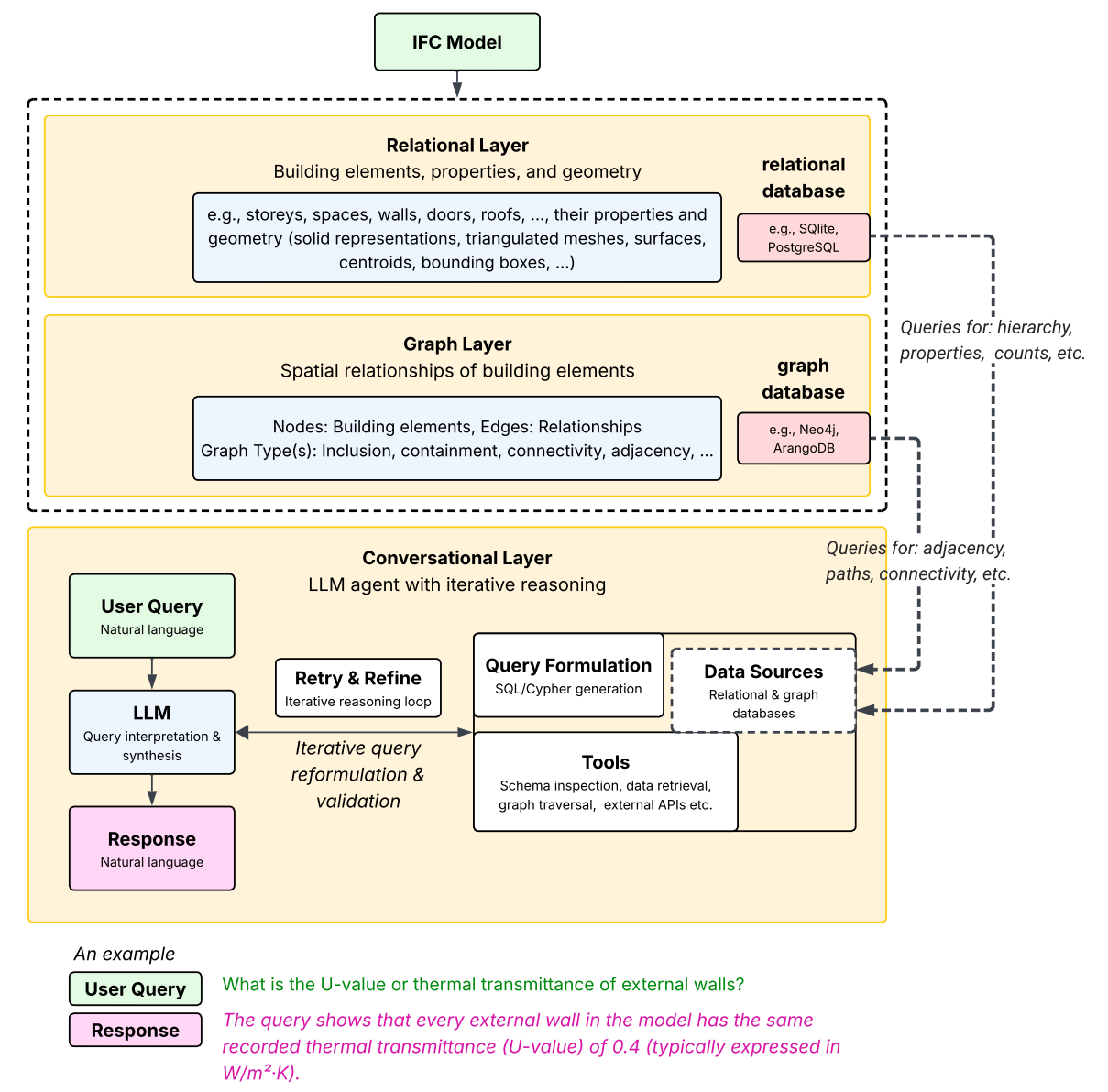}
    \caption{The three-layer architecture of IfcLLM. \revision{The relational and graph layers are named after the category of database each employs.}}
    \label{fig:overview}
\end{figure}

IfcLLM consists of three conceptual layers: (i) a relational layer, (ii) a graph layer, and (iii) a conversational layer, as illustrated in Figure~\ref{fig:overview}. \revision{The relational and graph layers are named after the category of database they employ, relational and graph respectively.} These layers provide complementary views of the same building model and work together to support natural language querying. The relational layer provides a structured representation of the IFC model at the element level. It captures building elements, their attributes, semantics, and geometry. The graph layer complements this by representing spatial relationships between elements. The conversational layer integrates these structured representations with an LLM agent, which interprets user questions, queries appropriate representations, and generates the final response. Next, we discuss each layer in detail.

\subsubsection{Relational Layer}

At this layer, the framework organizes building information around entities such as storeys, spaces, walls, doors, windows, slabs, columns, beams, roofs, and other relevant components. For each entity, the representation retains semantics and attributes such as names, descriptions, types, elevations, and properties, together with geometric information needed for downstream reasoning. The framework particularly preserves the hierarchical organization of the model, i.e., building>storey>element\{property\}. A conceptual data model for this layer is provided in Table \ref{table:ifcclasses}.

\begin{table}
\caption{A conceptual data model for the relational layer.}
\label{table:ifcclasses}
\centering
\footnotesize
\begin{tabular}{p{6cm} p{8cm}}
\toprule
\textbf{Concept} & \textbf{Attributes} \\
\midrule

\textbf{\underline{Building}} \\
Building entity 
& Identifier, name, description, general attributes \\

\textbf{\underline{Spatial hierarchy}} \\
Spatial unit (e.g., storey)
& Identifier, name, type, spatial position or level, parent reference, general attributes \\

\textbf{\underline{Element}} \\
Building element (e.g., room, wall, door, beam, column, etc.)
& Identifier, name, type, associated spatial unit, general attributes, simplified geometry, full geometry \\

\textbf{\underline{Property}} \\
Property set
& Associated element, property name, property value, value type, general attributes \\

\bottomrule
\end{tabular}
\end{table}

A key design choice in IfcLLM is to represent geometry at two complementary levels. The first is a simplified geometric representation, comprising geometric primitives such as centroids and axis-aligned bounding boxes. These can be precomputed and stored ahead of time, so that tasks such as distance estimation can be served immediately without incurring the cost of deriving such primitives on demand during LLM inference. The second is a full geometric representation, preserving complete mesh geometry for tasks such as visualization or more complex spatial analysis. This two-level design allows the framework to remain computationally practical while retaining the option for more detailed geometric reasoning when needed. The specific storage format chosen for each level is an implementation decision.

\subsubsection{Graph Layer}
\label{sec:graph-layer}
\begin{figure}
    \centering
    \includegraphics[width=1\linewidth]{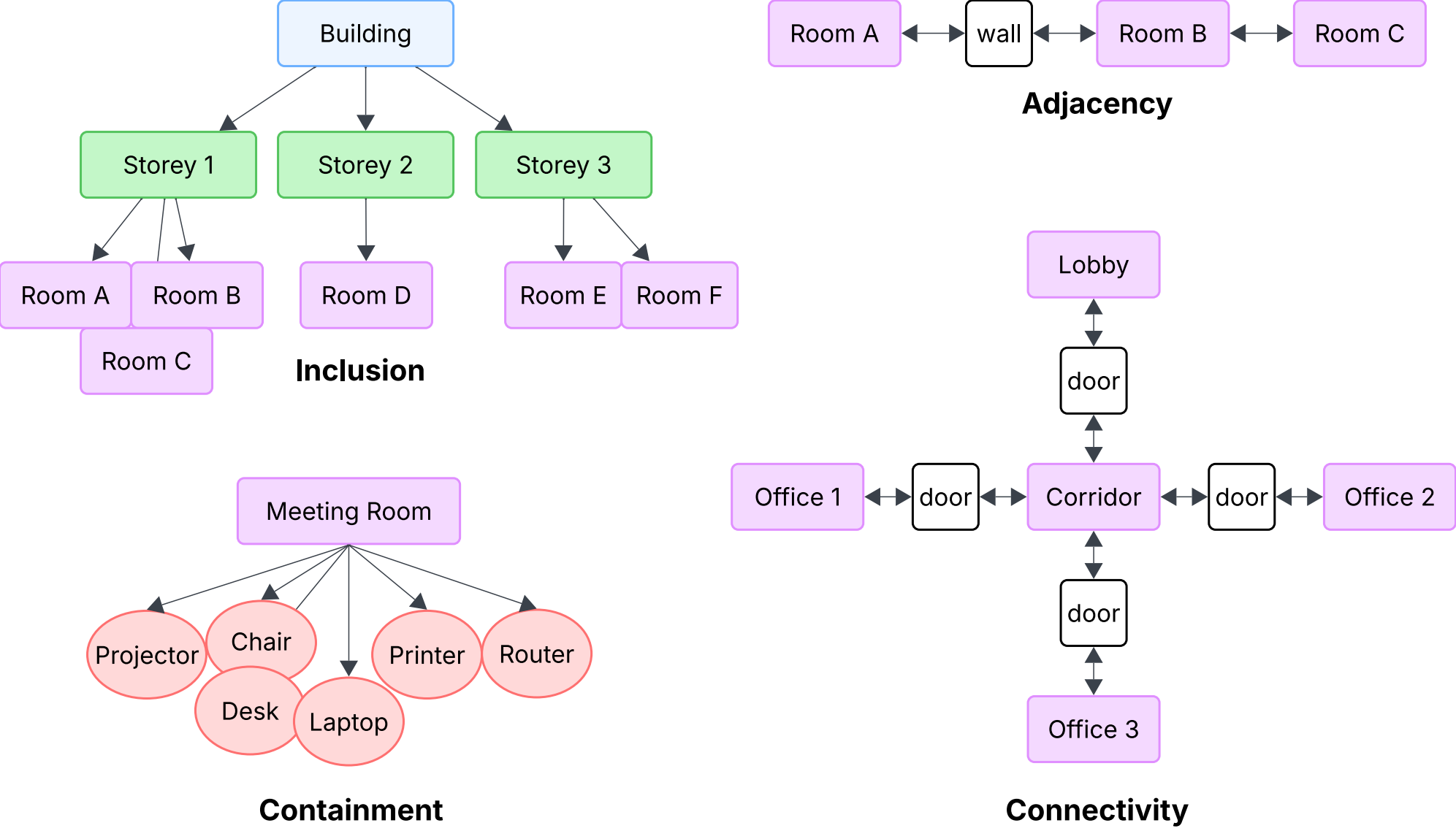}
    \caption{Example graph types for encoding spatial relationships between building elements. Note: These graphs and their definitions are illustrative and not exhaustive.}
    \label{fig:graphs}
\end{figure}

A defining characteristic of the graph layer is that it is task-oriented. Rather than representing all required relationships in IFC model as a single graph, the framework allows multiple graph types to be derived from the same IFC source, each capturing semantic or spatial relationships between selectively scoped building elements relevant to a specific application. This selective construction avoids large and dense graphs, reduces preprocessing and query overhead, and keeps each graph focused on its intended purpose. The semantic relations may refer to links between building elements without dealing with their geometry. The spatial relationships may capture topological properties such as inclusion, containment, adjacency, and connectivity (Figure \ref{fig:graphs}); directional properties such as left, right, above, and below; and metric properties such as proximity and distance. The graph layer acts as a complementary view to the relational layer, providing an interface for reasoning semantic and spatial relationships.

Numerous approaches have been proposed for deriving spatial graphs from IFC data \citep{liu2021indoor,diakite2022ifc2indoorgml,zhu2023ifc,zhu2025semantics}. The framework does not prescribe a specific spatial graph construction algorithm; any method that produces nodes and edges consistent with the graph layer's interface can be used. Our IfcLLM implementation, discussed later in Section \ref{sec:evaluation-setup}, utilizes a geometry-based strategy that uses axis-aligned bounding boxes to infer adjacency between selectively scoped building elements and create a connectivity graph.

\subsubsection{Conversational Layer}

\begin{algorithm}
\singlespacing
\scriptsize
\caption{An LLM-orchestrated algorithm for querying over relational and graph representations of IFC data}
\label{alg:ai-workflow}
\begin{algorithmic}[1]
\Require User query $q$ (text); LLM; relational DB $R$; graph DB $G$; prompts $P_{\text{sys}}, P_{\text{mid}}$.
\Ensure Natural language answer $a$.
\State Set $\textit{max\_iters} \leftarrow K$ (e.g., $K=5$), $\textit{iter} \leftarrow 0$
\State Initialize result store $\mathcal{D} \leftarrow \emptyset$ \Comment{$\mathcal{D}$ accumulates queries and their results}

\Statex
\State \textbf{Step 1: Prepare database summaries}
\State $S_R \leftarrow \textsc{SummarizeSchema}(R)$ \Comment{tables, columns, small examples}
\State $S_G \leftarrow \textsc{SummarizeGraph}(G)$ \Comment{node labels/properties, relationship types}
\State $\textit{context}_0 \leftarrow (q, S_R, S_G, P_{\text{sys}})$

\Statex
\State \textbf{Step 2: Decide actions}
\State $d \leftarrow \textsc{LLM}(\textit{context}_0)$ \Comment{structured decision output}
\If{$d.\textit{direct\_answer} = \textbf{true}$}
    \State \textbf{goto} Step 5
\EndIf
\State $Q_R \leftarrow d.\textit{sql\_queries}$; \quad $Q_G \leftarrow d.\textit{cypher\_queries}$

\Statex
\State \textbf{Step 3: Execute initial queries}
\If{$Q_R \neq \emptyset$}
    \State $r_R \leftarrow \textsc{ExecuteSQL}(R, Q_R)$
    \State $\mathcal{D} \leftarrow \mathcal{D} \cup \{(Q_R, r_R)\}$
\EndIf
\If{$Q_G \neq \emptyset$}
    \State $r_G \leftarrow \textsc{ExecuteCypher}(G, Q_G)$
    \State $\mathcal{D} \leftarrow \mathcal{D} \cup \{(Q_G, r_G)\}$
\EndIf

\Statex
\State \textbf{Step 4: Iterative completeness loop}
\While{$\textit{iter} < \textit{max\_iters}$}
    \State $\textit{iter} \leftarrow \textit{iter} + 1$
    \State $\textit{context}_{\textit{iter}} \leftarrow (q, S_R, S_G, \mathcal{D}, P_{\text{mid}})$
    \State $u \leftarrow \textsc{LLM}(\textit{context}_{\textit{iter}})$ \Comment{structured update output}
    \If{$u.\textit{analysis\_complete} = \textbf{true}$}
        \State \textbf{break}
    \EndIf
    \State $Q_R \leftarrow u.\textit{more\_sql\_queries}$; \quad $Q_G \leftarrow u.\textit{more\_cypher\_queries}$
    \If{$Q_R \neq \emptyset$}
        \State $r_R \leftarrow \textsc{ExecuteSQL}(R, Q_R)$
        \State $\mathcal{D} \leftarrow \mathcal{D} \cup \{(Q_R, r_R)\}$
    \EndIf
    \If{$Q_G \neq \emptyset$}
        \State $r_G \leftarrow \textsc{ExecuteCypher}(G, Q_G)$
        \State $\mathcal{D} \leftarrow \mathcal{D} \cup \{(Q_G, r_G)\}$
    \EndIf
\EndWhile

\Statex
\State \textbf{Step 5: Generate final answer}
\State $\textit{context}_{\text{final}} \leftarrow (q, \mathcal{D})$
\State $a \leftarrow \textsc{LLM}(\textit{context}_{\text{final}})$
\State \Return $a$
\end{algorithmic}
\end{algorithm}

At this layer, an LLM operates as an agent with access to multiple tools for structured data retrieval from the above constructed relational and graph representations of IFC data. Given a user query, the agent first determines whether the answer can be produced directly from general knowledge or requires retrieval from the backends. This decision is guided by structured schema descriptions of both representations and explicit routing guidelines embedded in the agent's system prompt: for instance, attribute retrieval, aggregation, and hierarchy queries are directed to the relational layer, while spatial traversal, adjacency, and shortest-path queries are directed to the graph layer. When retrieval is needed, the agent formulates structured queries against the backends, receives the results, and uses them to guide subsequent steps.

The framework adopts a retry-and-refine strategy, formalized in Algorithm \ref{alg:ai-workflow}, to handle queries that involve multiple constraints or require combining data from both representations. On each iteration, the agent receives the original user query, schema summaries of both representations, and the accumulated history of previously executed queries and their results. Based on this context, it evaluates whether the available evidence is sufficient to answer the query. If not, it generates additional queries against one or both backends, incorporating the intermediate results into the next iteration. If a query fails due to a syntax error or schema mismatch, the agent reformulates it using the error feedback as additional context. This loop continues until the agent determines the analysis is complete or a predefined iteration limit is reached, at which point the accumulated evidence is used to synthesize the final natural language response. The conversational layer therefore acts as the reasoning and control component of the framework.

The framework is designed to be implementation-agnostic. The relational layer can be backed by any SQL-compatible system, from lightweight embedded databases for local deployments to full relational database servers or spatially extended variants that support native geometry types and spatial functions. The graph layer can be realized using different graph databases or different graph construction methods. The conversational layer can integrate different LLMs, whether open-weight models deployed locally or proprietary API-based services, and may employ different reasoning strategies depending on the capabilities of the chosen model.

\section{Implementation and Evaluation Setup}
\label{sec:evaluation-setup}

To evaluate the proposed framework, we implement an end-to-end system that operationalizes the relational, graph, and conversational layers described in Section~\ref{sec:framework}. This section outlines the key implementation choices, and evaluation models, queries and criteria.

The implementation is a functional prototype scoped to demonstrate the feasibility of the proposed framework. The relational layer covers a representative subset of building concepts: building, storey, beam, ceiling, column, door, railing, ramp, roof, room (space), slab, stair, transport, wall, and window. The graph layer is further restricted to room (space), door, and stair, consistent with a set of spatial queries evaluated in this work (discussed later in Section~\ref{sec:evaluation-queries}). A key dependency between the two layers is that the relational layer precomputes geometric primitives during IFC transformation for the elements included in the graph layer, which are subsequently reused to construct a connectivity graph.

\subsection{IFC Data Transformation}
\label{sec:design-choices}
\begin{figure}
    \centering
    \includegraphics[width=\textwidth]{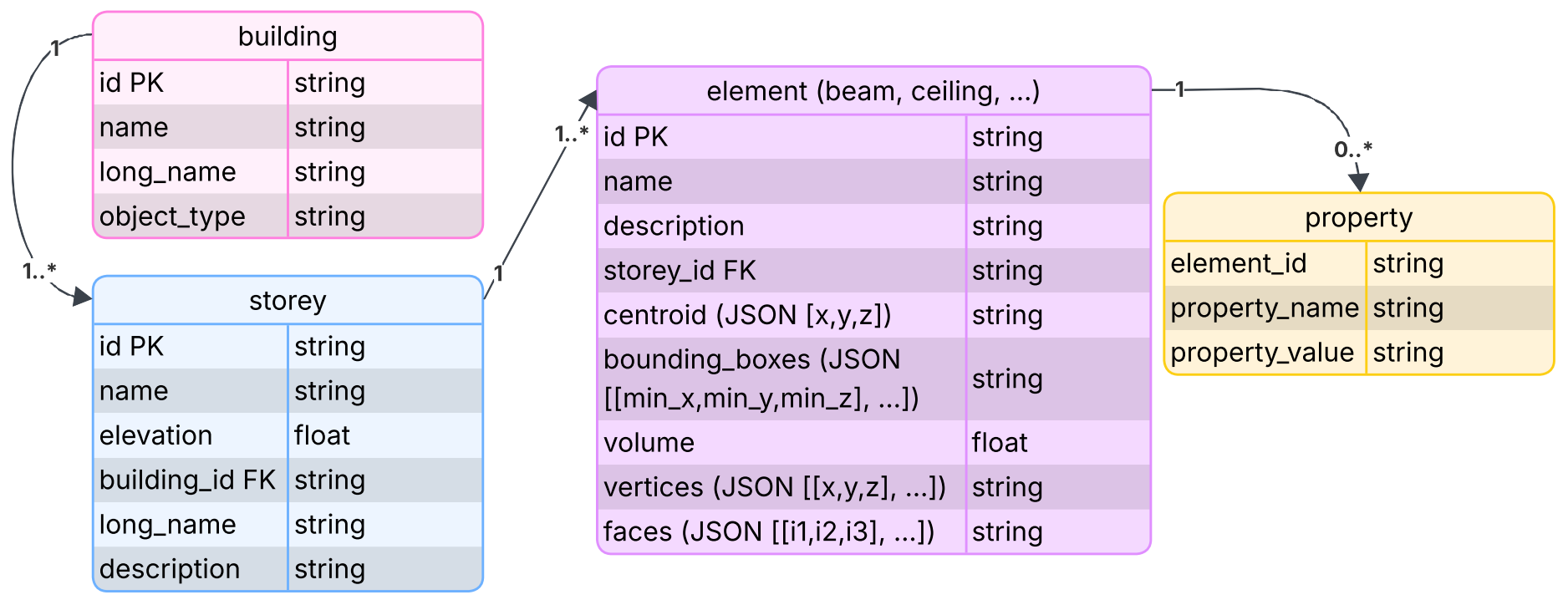}
    \caption{UML class diagram of the implemented relational schema.}
    \label{fig:relational-schema}
\end{figure}

Selected building elements, their properties, and geometric information are extracted from IFC files using IfcOpenShell and its geometric engine \citep{ifcopenshell}. The extracted data are stored in a SQLite database. SQLite is chosen for its simplicity, excellent performance, zero-configuration deployment, and broad compatibility, making it well suited for prototyping without requiring a dedicated database server. For all IFC models evaluated in this study, a total of 16 relational tables were generated: one table for each building concept, and a property table\footnote{We adopt an entity-attribute-value pattern, limiting the schema to a single table regardless of the property sets available across IFC models and keeping attribute-based querying such as filtering and aggregation straightforward.}. The implemented relational schema is illustrated in Figure \ref{fig:relational-schema}, and relational table names and their corresponding IFC entity classes are listed in Table \ref{tab:ifc-mapping}.

In this implementation, centroids and bounding boxes are the selected simplified geometric representations, used for distance estimation and adjacency detection respectively (the latter when constructing the connectivity graph considered in this work). Since adjacency and distance computations in this prototype are performed in application code rather than within SQL, native spatial data types, provided by extensions such as SpatiaLite, are not used. Incorporating such spatial extensions to support native spatial functions is identified as future work (Section~\ref{sec:discussion}). These simplified representations are stored in a compact JSON array format, with centroids represented as \texttt{[x, y, z]} and bounding boxes as \texttt{[[min\_x, min\_y, min\_z], [max\_x, max\_y, max\_z]]}.

Full geometric representations are similarly stored as JSON arrays of vertices and faces. SQLite's built-in JSON functions allow convenient extraction of this data whenever required. To establish hierarchical relationships, storey elevations are first extracted, and each element is assigned to a corresponding storey based on its placement in the building coordinate system. This placement is determined by recursively traversing the IFC object hierarchy and accumulating relative placements. For elements spanning multiple storeys, a single representative storey is assigned to simplify querying.

\begin{table}
\centering
\footnotesize
\caption{\secondrevision{Relational table names, corresponding IFC entity
classes, and standardized property sets observed across the three evaluated models.}}
\label{tab:ifc-mapping}

\begin{tabular}{lll}
\toprule
\textbf{Table} &
\textbf{IFC Entity Class(es)} &
\textbf{\secondrevision{Standardized Property Set(s)\textsuperscript{a}}} \\
\midrule
building  & IfcBuilding
          & Pset\_BuildingCommon \\
\midrule
storey    & IfcBuildingStorey
          & None observed\textsuperscript{b} \\
\midrule
beam      & IfcBeam
          & Pset\_BeamCommon \\
ceiling   & IfcCovering (CEILING)
          & Pset\_CoveringCommon \\
column    & IfcColumn
          & Pset\_ColumnCommon \\
door      & IfcDoor
          & Pset\_DoorCommon \\
railing   & IfcRailing
          & Pset\_RailingCommon \\
ramp      & IfcRamp
          & No instances\textsuperscript{c} \\
roof      & IfcRoof
          & Pset\_RoofCommon \\
room      & IfcSpace
          & Pset\_SpaceCommon, Pset\_SpaceThermalRequirements \\
slab      & IfcSlab
          & Pset\_SlabCommon \\
stair     & IfcStair, IfcStairFlight
          & Pset\_StairCommon, Pset\_StairFlightCommon \\
transport & IfcTransportElement
          & No instances\textsuperscript{c} \\
wall      & IfcWall, IfcWallStandardCase
          & Pset\_WallCommon \\
window    & IfcWindow
          & Pset\_WindowCommon \\
\midrule
property  & IfcPropertySet
          & All extracted property sets\textsuperscript{a} \\
\bottomrule
\end{tabular}

\vspace{2pt}
\begin{minipage}{0.96\linewidth}
\footnotesize
\secondrevision{\textsuperscript{a} Property extraction is based on IfcPropertySet. The
column lists the standardized property sets observed in at least one of
the evaluated models; exporter-specific sets such as PSet\_Revit\_* are
also stored.}

\secondrevision{\textsuperscript{b} Storeys occurred in all three models. No property sets
were associated with storeys in one model, while the other two contained
only exporter-specific PSet\_Revit\_* sets.}

\secondrevision{\textsuperscript{c} No instance of this entity class occurred in any of
the three evaluated models.}
\end{minipage}
\end{table}

Next, for the graph layer we generate a connectivity graph over rooms (spaces), doors, and stairs (hereinafter also referred to as \textit{navigable elements}). We perform adjacency checks between the axis-aligned bounding boxes of the navigable elements, as per Algorithm~\ref{algo:adjacency_detection}. When adjacency is detected, the corresponding elements are connected, resulting in an incrementally constructed graph. To improve connectivity, we apply a lightweight adjustment to door elements. Since doors are typically embedded within walls, their bounding boxes may not always intersect adjacent spaces, leading to missing connections. To address this, door bounding boxes are expanded along the wall thickness (by $x$ units). This correction improves the connectivity representation without introducing significant computational overhead. We use \textit{Neo4j} \citep{webber2012programmatic} to store the resulting graph, as it is a widely adopted system with efficient support for property graphs and graph traversal operations \citep{zhu2023ifc}.

\begin{algorithm}[t]
\scriptsize
\caption{Connectivity graph generation based on axis-aligned bounding boxes}
\begin{algorithmic}[1]
\Require{Bounding boxes $B_1 = (\mathbf{b}_1^{min}, \mathbf{b}_1^{max})$, $B_2 = (\mathbf{b}_2^{min}, \mathbf{b}_2^{max})$ where $\mathbf{b}_i^{min}, \mathbf{b}_i^{max} \in \mathbb{R}^3$}; Tolerance $\epsilon \geq 0$; containment flag $\gamma \in \{\text{true}, \text{false}\}$
\Ensure{Boolean adjacency result}
\Function{BoundingBoxesAreAdjacent}{$B_1$, $B_2$, $\epsilon$, $\gamma$}
    \State $(\mathbf{m}_1, \mathbf{M}_1) \gets B_1$ \Comment{Extract min/max vectors}
    \State $(\mathbf{m}_2, \mathbf{M}_2) \gets B_2$
    
    \If{$\gamma = \text{true}$} \Comment{Check intersection and containment}
        \State $\text{intersects} \gets \bigwedge_{i=1}^{3} (M_{1,i} \geq m_{2,i} - \epsilon \land m_{1,i} \leq M_{2,i} + \epsilon)$
        \If{$\text{intersects}$}
            \State \Return $\text{intersect:true}$
        \EndIf
        
        \State $C_1 \gets \bigwedge_{i=1}^{3} (m_{1,i} \leq m_{2,i} + \epsilon \land M_{1,i} \geq M_{2,i} - \epsilon)$
        \State $C_2 \gets \bigwedge_{i=1}^{3} (m_{2,i} \leq m_{1,i} + \epsilon \land M_{2,i} \geq M_{1,i} - \epsilon)$
        \If{$C_1 \lor C_2$}
            \State \Return $\text{containment:true}$
        \EndIf
    \EndIf
    
    \For{$k \gets 1$ \textbf{to} $3$} \Comment{Test face adjacency across all axes}
        \State $\text{touching}_k \gets (|M_{1,k} - m_{2,k}| < \epsilon) \lor (|M_{2,k} - m_{1,k}| < \epsilon)$
        \If{$\text{touching}_k$}
            \State $\text{overlap} \gets \bigwedge_{j \neq k} (m_{1,j} < M_{2,j} - \epsilon \land M_{1,j} > m_{2,j} + \epsilon)$
            \If{$\text{overlap}$}
                \State \Return $\text{face-adjacency:true}$
            \EndIf
        \EndIf
    \EndFor
    
    \State \Return $\text{adjacency:false}$
\EndFunction
\end{algorithmic}
\label{algo:adjacency_detection}
\end{algorithm}

\subsection{LLM Workflow}
We use LangGraph \citep{langgraph2025} to implement the LLM workflow described in Algorithm~\ref{alg:ai-workflow}. The retry-and-refine mechanism permits a maximum of five iterations in all experiments. To ensure a controlled comparison, the workflow is kept fixed across models, including the data representations, system and intermediate prompts, tool interfaces, and reasoning procedure. \secondrevision{For each experiment, the selected LLM is used consistently throughout query generation, iterative refinement, and final answer synthesis. This ensures that comparisons across LLMs isolate differences in foundation-model behaviour under a common workflow.}

\revision{As IfcLLM is intended for fully local deployment, we focus primarily on open-weight LLMs. We evaluate four open-weight LLMs from different families and parameter scales: GPT-OSS 120B and GPT-OSS 20B \citep{agarwal2025gpt}, Gemma 4 31B \citep{team2026gemma}, and Llama 3.3 70B \citep{grattafiori2024Llama}. GPT-5.6 Luna \citep{openai2026gpt56} is additionally included as a proprietary baseline to provide a frontier-model reference for interpreting the performance of the open-weight models. The GPT-OSS models enable a within-family comparison across model scales, while Gemma 4 31B and Llama 3.3 70B broaden the evaluation across independent model families. Although larger open-weight models, including those in the trillion-parameter range, are available, we restrict our selection to models that can be deployed for inference on a single 80~GB GPU, consistent with the local-deployment objective.}

We provide the system prompt template in \ref{system-prompt} and the intermediate prompt template in \ref{intermediate-prompt}. LLM inference is performed using the OpenAI API library\footnote{https://pypi.org/project/openai/} which is natively supported by several LLM inference providers, including locally hosted OpenAI-compatible endpoints. \revision{For all LLMs, the temperature was set to 0.1 and top\_p to 1.0, while the maximum output token limit was set to the LLM's default.}

\subsection{IFC Building Models}
\begin{figure}
    \centering
    % Left column (2 stacked images)
    \begin{minipage}[t]{0.49\textwidth}
        \centering
        \includegraphics[width=0.9\linewidth]{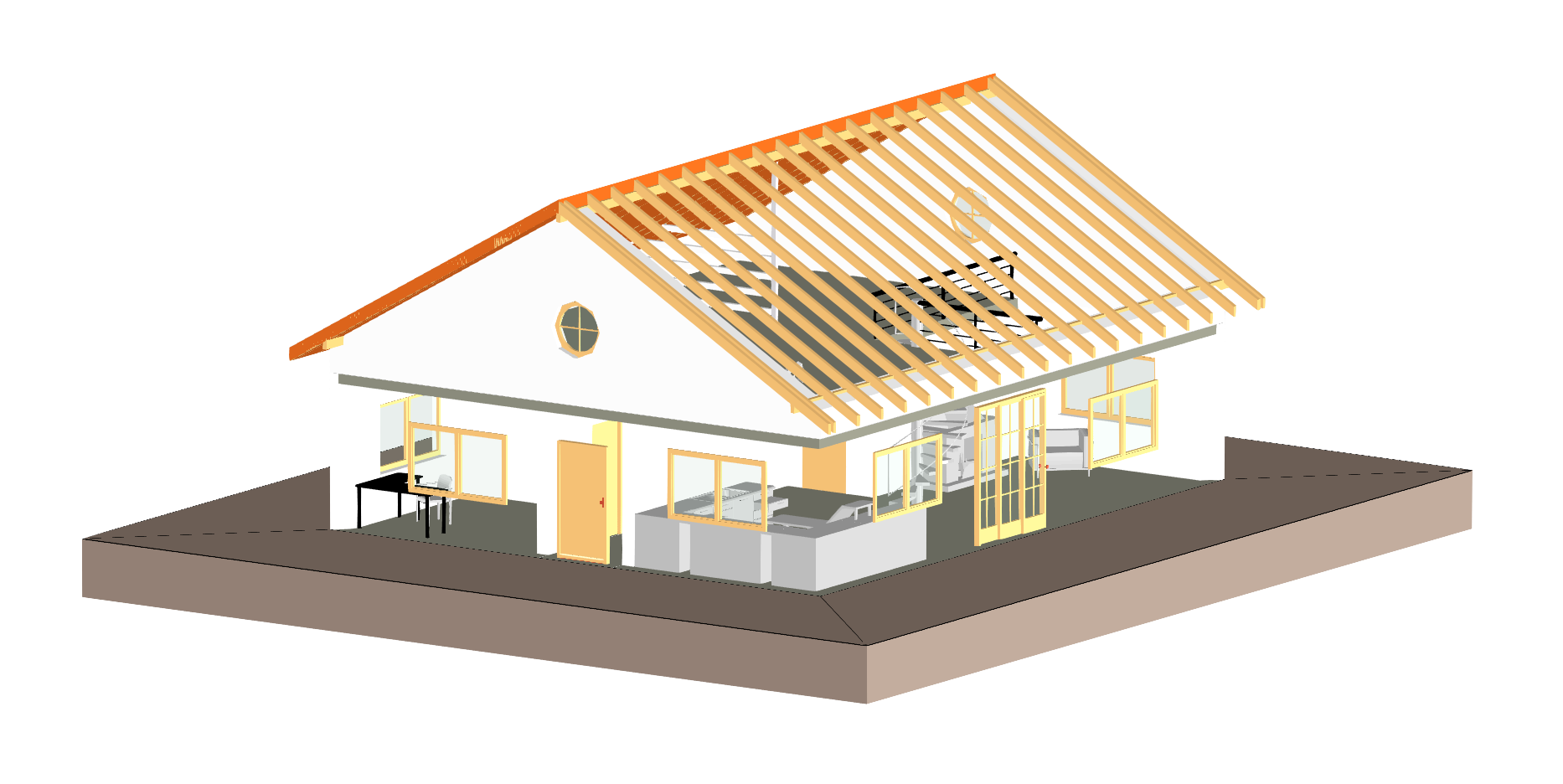}
            \subcaption*{(a) IFC model}
        \hfill
    \end{minipage}
    \begin{minipage}[t]{0.49\textwidth}
            \centering
            \includegraphics[width=0.85\linewidth]{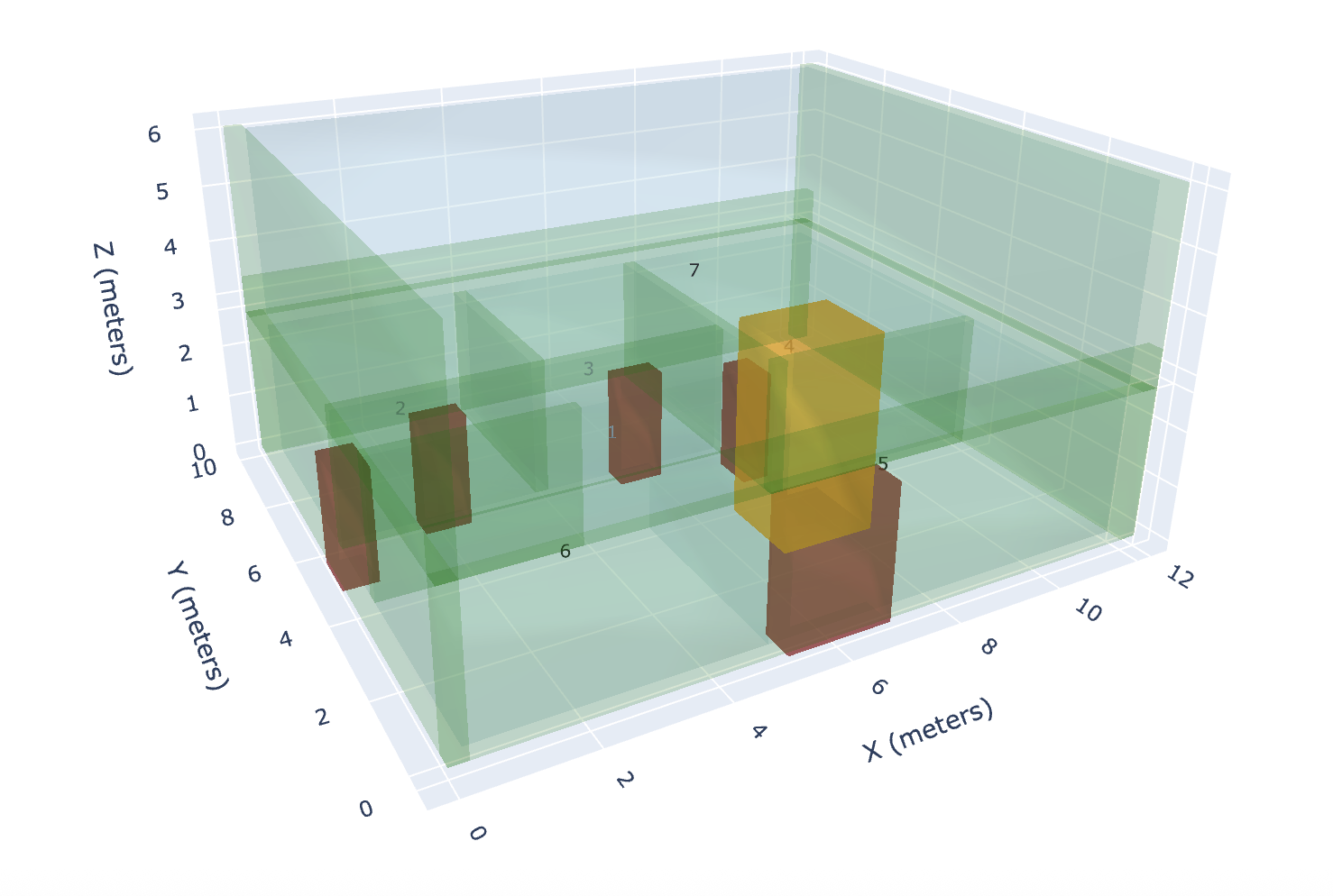}
            \subcaption*{(b) Bounding box plot}
    \end{minipage}
    
    \vspace{2em}
    
    \begin{minipage}[t]{\textwidth}
    \centering
    \hspace*{-2.5cm}
    \includegraphics[height=0.75\textheight,keepaspectratio]{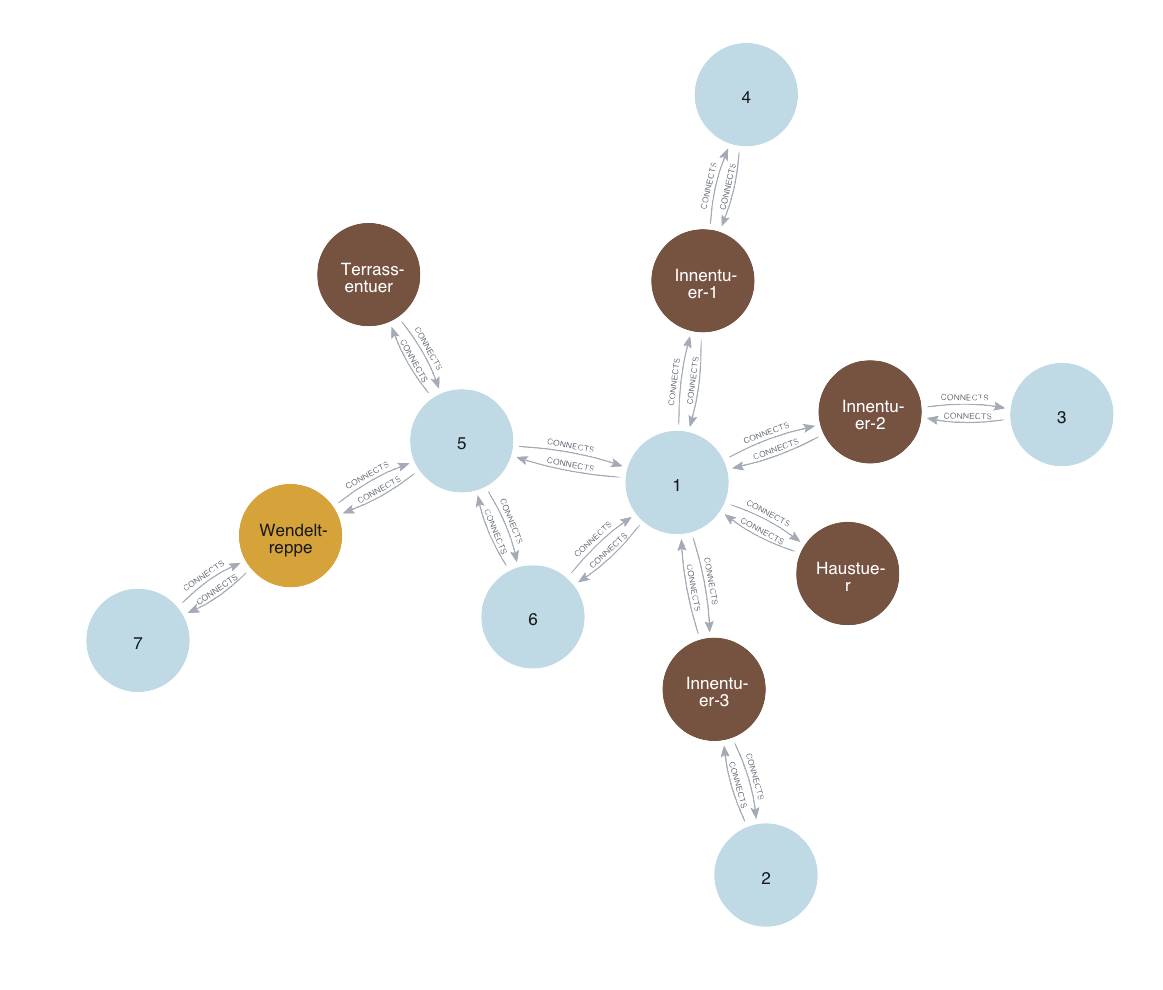}
        \subcaption*{(c) Connectivity graph}
    \end{minipage}

    \caption{A simple 2-storey house (Building B1) with a single stair and multiple doors. The IFC model used is the ``FZK Haus" provided by Karlsruhe Institute of Technology (KIT) (retrieved from https://openifcmodel.cs.auckland.ac.nz). Building elements in (b) and nodes in (c): \legendcircle{roomcolor} Rooms, \legendcircle{staircolor} Stair, \legendcircle{doorcolor} Doors.}
    \label{fig:hauz-ifc}
\end{figure}

\begin{figure}
    \centering

    \vspace{-8em}

    \begin{minipage}[t]{0.48\textwidth}
        \centering
        \includegraphics[width=1\linewidth]{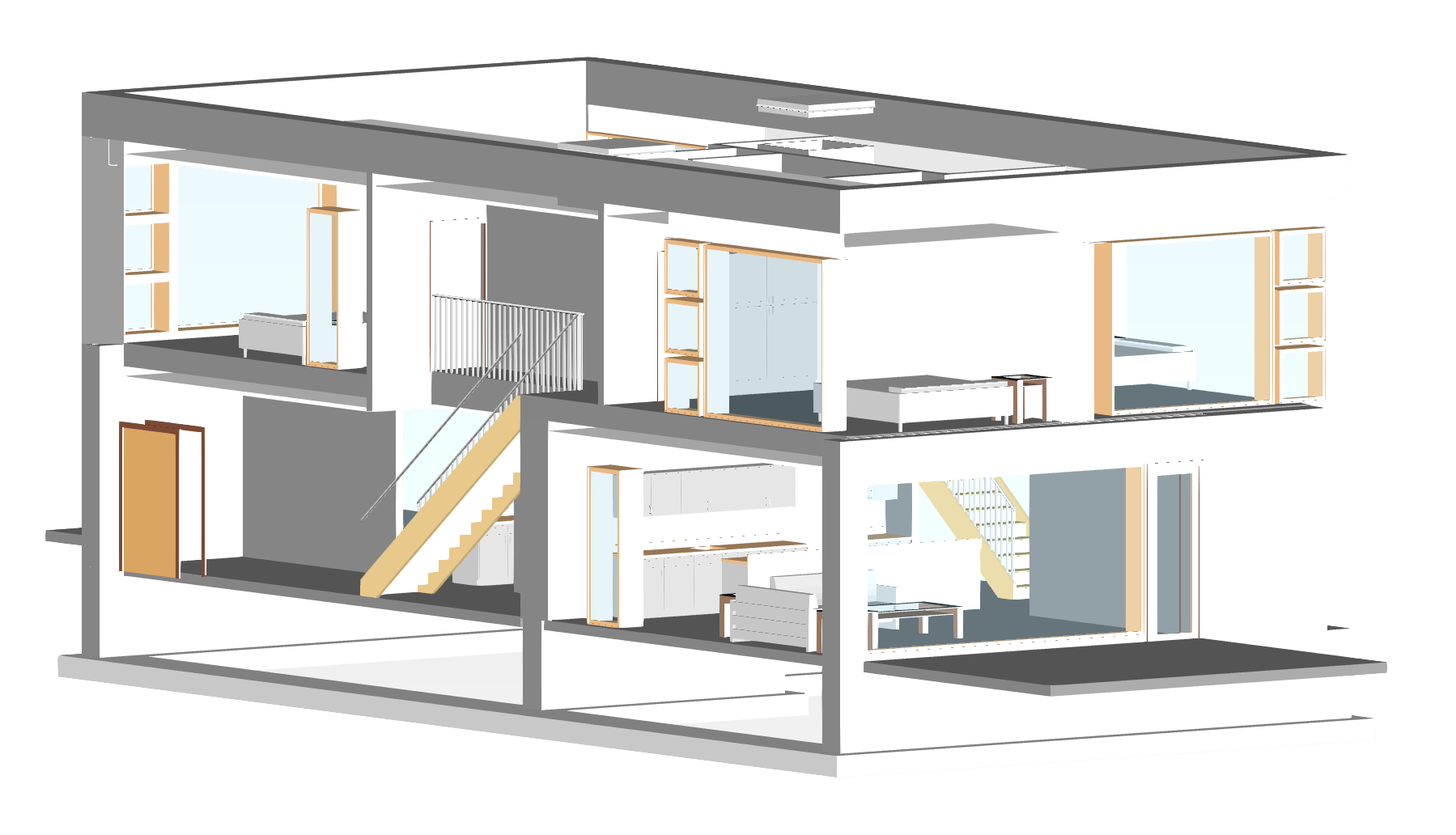}
        \subcaption*{(a) IFC model}
    \end{minipage}
    \hfill
    \begin{minipage}[t]{0.48\textwidth}
        \centering
        \includegraphics[width=1\linewidth]{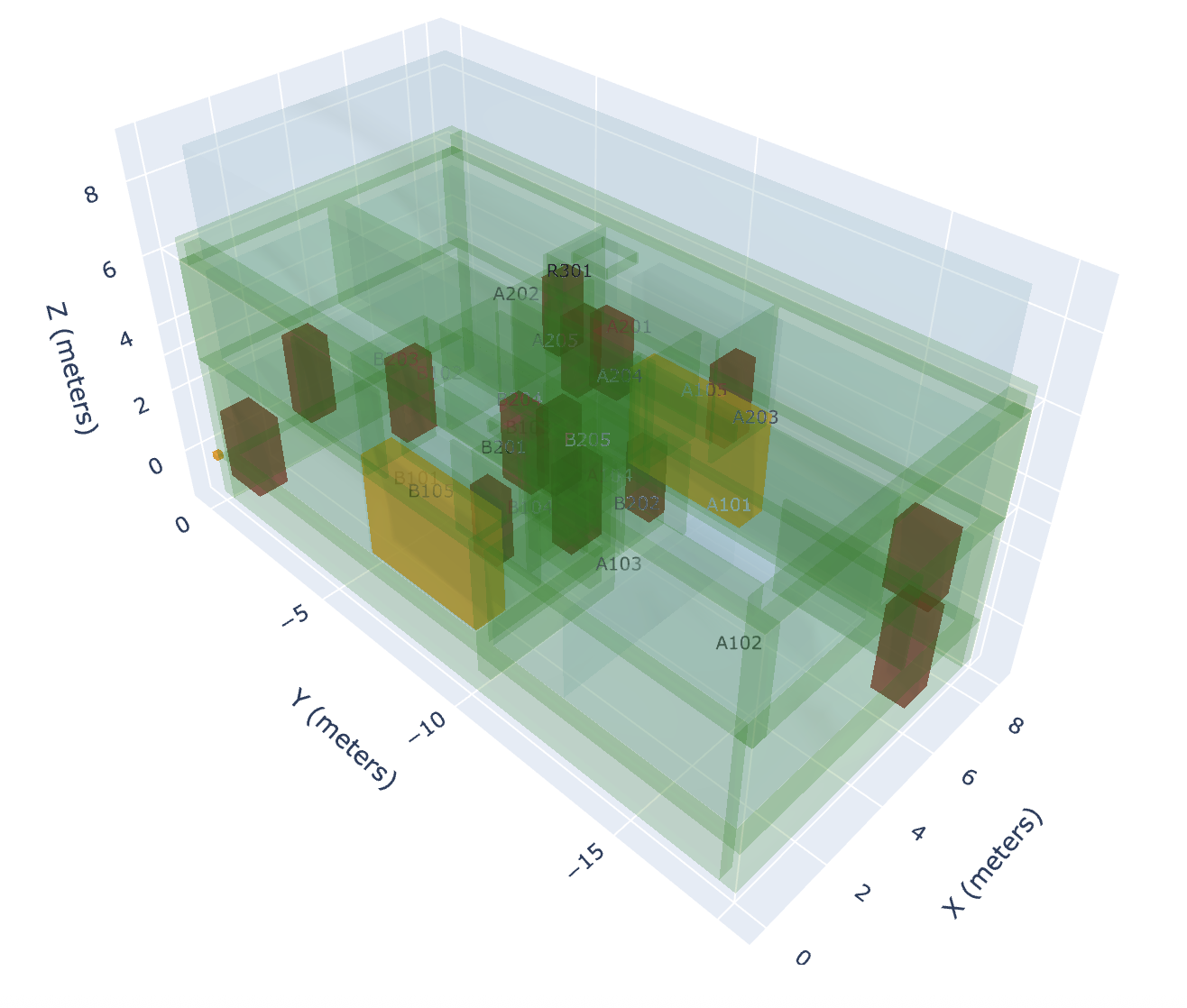}
        \subcaption*{(b) Bounding box plot}
    \end{minipage}

    \vspace{3em}

    \begin{minipage}[t]{\textwidth}
        \centering
        \hspace*{-4.65cm}
        \includegraphics[width=1.6\linewidth]{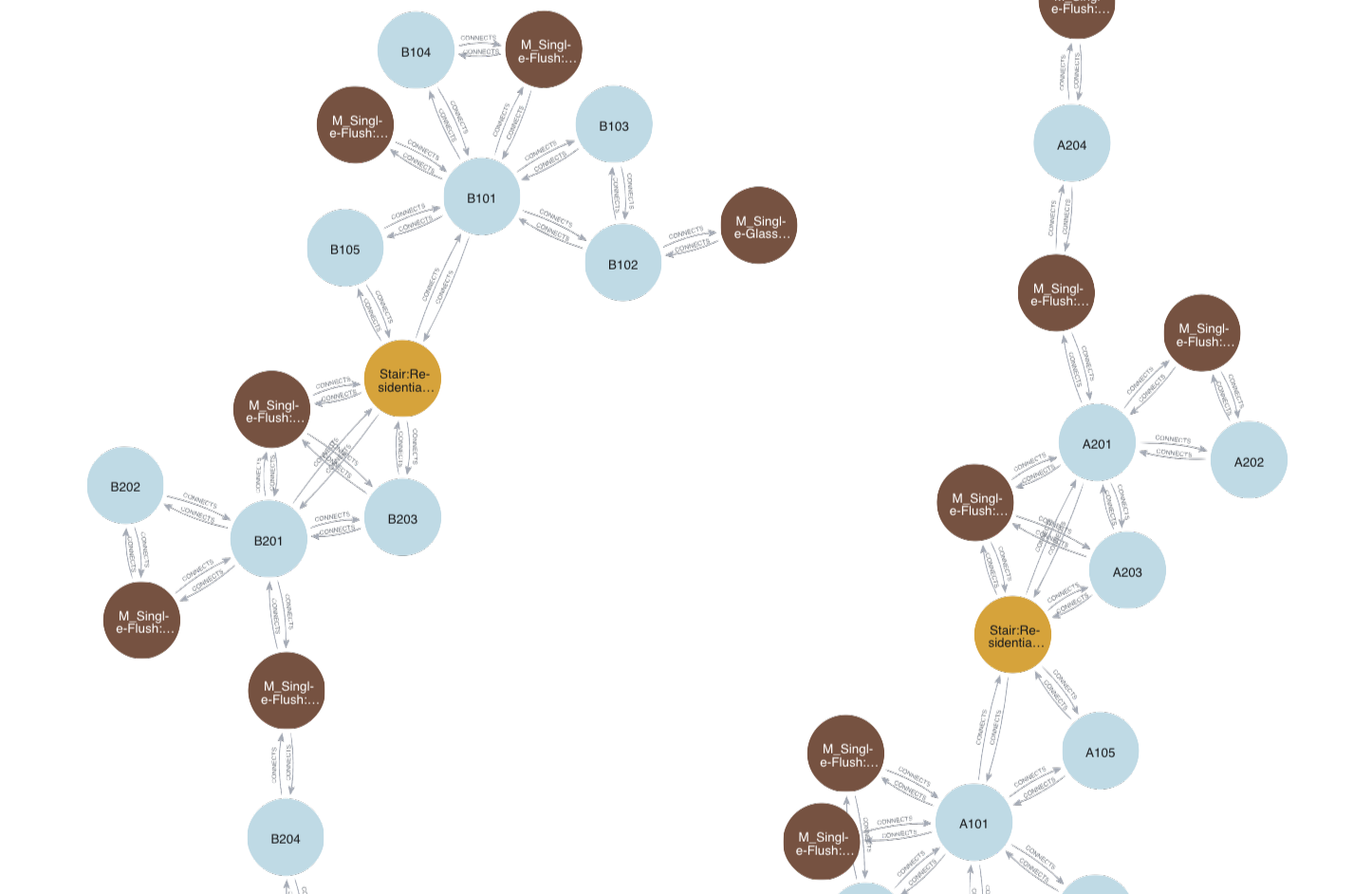}
        \subcaption*{(c) Connectivity graph (subgraph shown)}
    \end{minipage}

    \caption{A duplex building (Building B2) with two separate apartments, each having its own entrance and no internal connection. The IFC model used is the ``Duplex Apartment" provided by buildingSMART Alliance (https://nibs.org/). Building elements in (b) and nodes in (c): \legendcircle{roomcolor} Rooms, \legendcircle{staircolor} Stair, \legendcircle{doorcolor} Doors.}
    \label{fig:duplex-ifc}
\end{figure}

\begin{figure}
    \centering
    
    \begin{subfigure}{0.48\textwidth}
        \centering
        \includegraphics[width=1\linewidth]{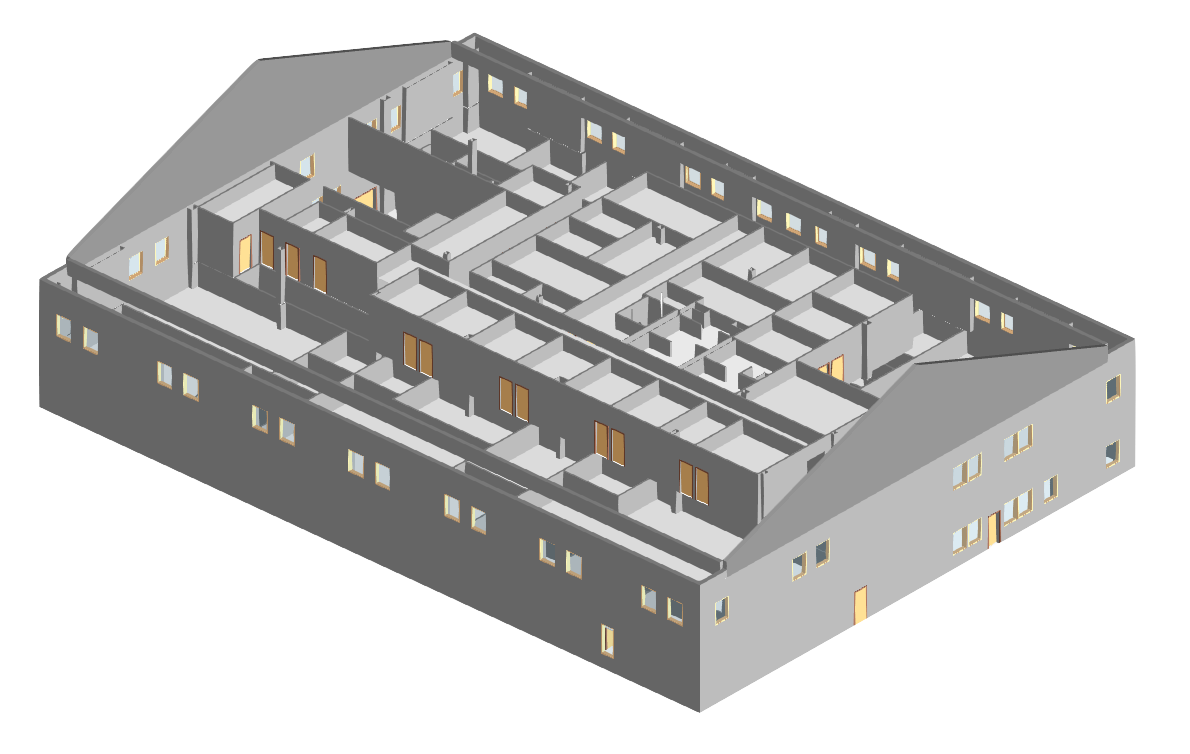}
        \caption{IFC model}
    \end{subfigure}
    \hfill
    \begin{subfigure}{0.48\textwidth}
        \centering
        \includegraphics[width=1\linewidth]{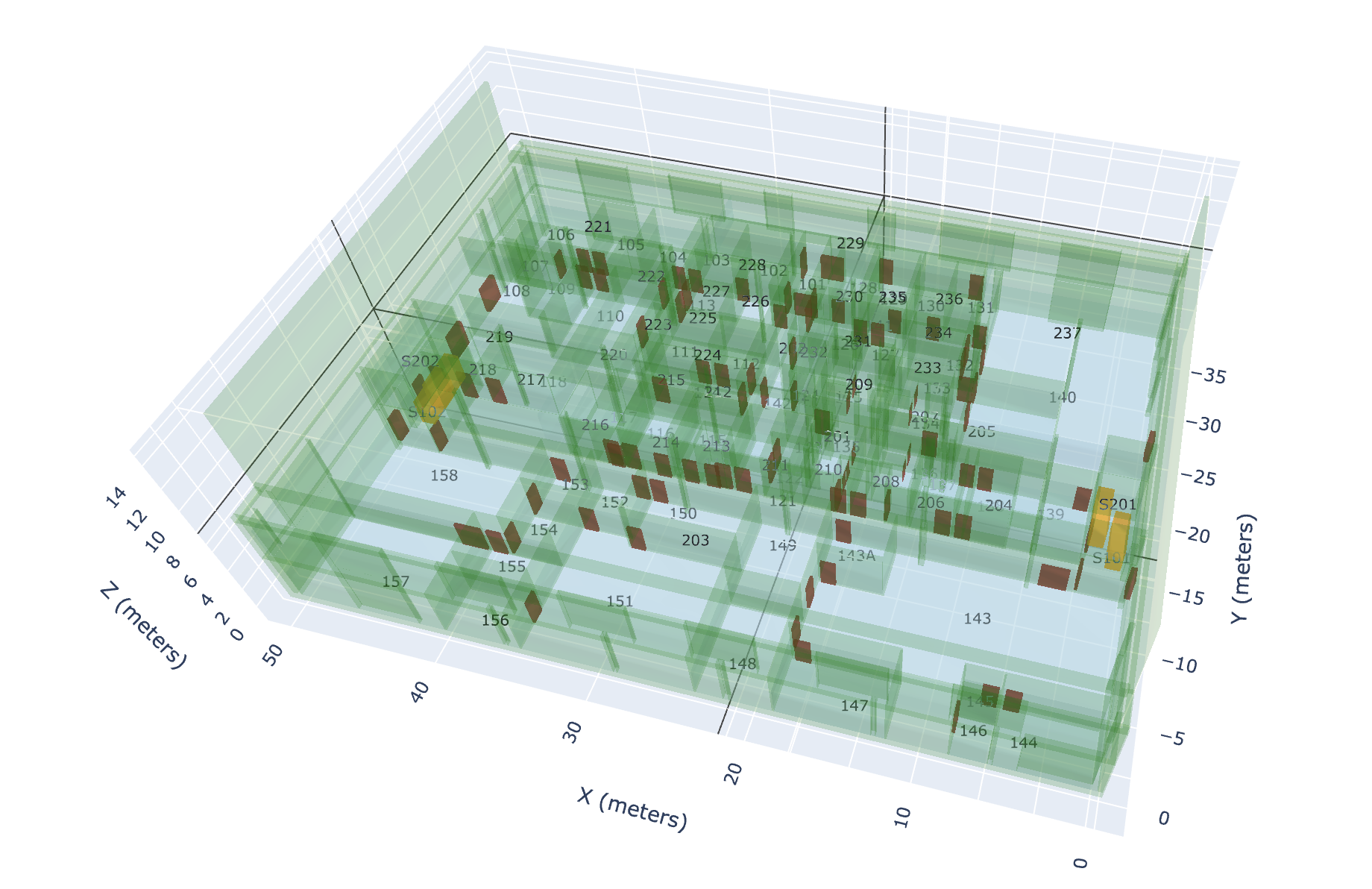}
        \caption{Bounding box plot}
    \end{subfigure}
\caption{An office building (Building B3). The IFC model is provided by buildingSMART Alliance (https://nibs.org/). Building elements in (b): \legendcircle{roomcolor} Rooms, \legendcircle{staircolor} Stair, \legendcircle{doorcolor} Doors.}
\label{fig:office-building}
\end{figure}

We utilize three openly available and frequently used IFC models for evaluation. \textit{Building B1} is a small two-storey house with an internal staircase, \textit{Building B2} is a duplex comprising two independent apartments with separate entrances and no internal connectivity between them, and \textit{Building B3} is an office building. Key statistics for these IFC models are provided in Table~\ref{tab:stats}. Figures~\ref{fig:hauz-ifc}--\ref{fig:office-building} show the IFC models and related representations. Buildings B1 and B2 are relatively small and are selected for ease of visualization. Building B2 serves a further purpose: the bounding boxes of certain rooms overlap, which offers a controlled case for examining the implemented adjacency detection method under non-trivial room geometries. Building B3, roughly an order of magnitude larger, is used to assess scalability on a model closer in size to real-world office buildings.

\begin{table}[t]
\footnotesize
    \caption{Key statistics of the IFC models used in the evaluation. }
    \label{tab:stats}
    \centering
    \begin{tabular}{c|c|c|c|c|c}
    \toprule
     \textbf{IFC Model} & \textbf{storey} & \textbf{rooms} & \textbf{doors} & \textbf{stairs} & \textbf{properties} (distinct)\\
      \midrule
       Building B1 & 2 & 7 & 5 & 1 & 321 (21)\\
       Building B2 & 4 & 21 & 14 & 2 & 12,478 (213)\\
       Building B3 & 3 & 99 & 102 & 4 & 78,758 (140)\\
       \bottomrule
    \end{tabular}
\end{table}

\subsection{Evaluation Queries}
\label{sec:evaluation-queries}
To evaluate the proposed framework, we constructed a set of natural language queries based on 30 representative scenarios reflecting common information needs when interacting with IFC-based BIM models (\ref{all-queries-motivation}). These scenarios capture realistic tasks encountered by AEC stakeholders and include both standalone queries and sequences of follow-up queries that arise during exploratory interactions. The query set is not intended to be exhaustive; rather, it is designed to assess the functional behavior of the framework, including its ability to interpret natural language inputs, select the appropriate backend(s), generate executable queries, and return meaningful natural language response.

Importantly, the queries represent a set of recurring query patterns that can be parameterized across different IFC models. For example, a query such as ``\textit{What is the fire rating of door ‘X’?}'' follows a general structure of retrieving a property for a given element (i.e., \textit{[property] of [element]}), while ``\textit{Which doors are exit doors?}'' reflects a filtering pattern over a class of elements based on a condition (i.e., \textit{[elements] satisfying [property constraint]}). Similarly, aggregation queries (e.g., counts per storey) and connectivity queries (e.g., path existence between spaces) follow consistent structural forms. By varying the element type, property, or constraints, these query templates can be directly adapted to different elements and contexts within a model, and reused across different IFC models without changing the underlying logic.

To support this objective, the queries are categorized as:

\begin{itemize}
\item \textit{SQL-based queries}, which primarily involve attribute retrieval, filtering, aggregation, and comparison of building elements.
\item \textit{Graph-based queries}, which primarily involve connectivity, reachability, and navigation between building elements.
\end{itemize}

This categorization reflects the hybrid design of the framework and allows us to evaluate whether the framework can correctly route queries to the appropriate backend and handle different query paradigms. Table~\ref{tab:scenarios} illustrates representative user scenarios that motivate the query set, while Table~\ref{tab:querytypes} provides example queries for each category. The complete set of scenarios is provided in \ref{all-queries-motivation}.

\begin{table}[t!]
\caption{Example user scenarios motivating evaluation queries.}
\label{tab:scenarios}
\centering
\footnotesize
\begin{tabular}{p{3cm} p{11cm}}
\toprule
\textbf{Stakeholder} & \textbf{Example Scenario} \\
\midrule
Facility Manager & A facility or campus operations manager might query the IFC model for the number of storeys in a building to estimate cleaning workloads, allocate maintenance staff, or plan security patrol coverage across floors. \\
Fire Safety Officer & A fire safety officer might query the IFC to identify which room has the most doors in order to ensure it is properly marked and managed for evacuation routes. \\
Energy Consultant & An energy consultant or HVAC engineer could query the IFC model for the U-value (thermal transmittance) of the external walls to assess the building’s energy performance and ensure compliance. \\
Security Planner & A facility manager or safety and security planner might query the IFC to find which rooms are directly connected to room “X” to plan access control or service routes. \\
BIM Analyst & A BIM analyst might retrieve the bounding box of room “X” from the IFC model to quickly approximate its spatial extent when performing spatial indexing, collision checks, or coarse spatial analysis. \\
\bottomrule
\end{tabular}
\end{table}

\begin{table}[t!]
\caption{Examples of \textit{SQL-based} and \textit{graph-based} queries.}
\label{tab:querytypes}
\centering
\footnotesize
\begin{tabular}{p{3cm} p{11cm}}
\toprule
\textbf{Query Type} & \textbf{Example Query} \\
\midrule
\multirow{4}{*}{SQL-based} & Which slabs or ceilings belong to the bottom-most floor? \\
 & Which doors are exit doors?\\
 & What is the fire rating of door ``X"? \\
 & What are the elevations of the ground and the top-most floor?\\
\midrule
\multirow{3}{*}{Graph-based} & Which rooms are directly connected to room ``X"? \\
 & Is there a navigable path from the “Lobby" to Room ``X"? \\
 & Which rooms are isolated (not reachable from any other room)? \\
\bottomrule
\end{tabular}
\end{table}

\subsection{Evaluation Criteria}
\label{evaluation-criteria}
We evaluate the framework using the following metrics. \textit{Accuracy} measures the percentage of queries for which the system returns the correct final answer. A response is counted as correct if the final natural language answer matches the information stored in the IFC model for that query. \revision{\textit{Token usage} measures the number of input and output tokens processed by the LLM. \textit{LLM calls} measures how many LLM invocations are required to produce an answer. \textit{SQL and Cypher query counts} measure the number of database queries issued to retrieve the required information. Token usage, LLM calls, and SQL/Cypher query counts are reported as means.}

We also evaluate the framework through an ablation study that isolates the impact of its core design choices. Accordingly, we evaluate five system variants: 

\begin{itemize}
    \item \textit{IfcLLM}: The hybrid system, combining relational and graph representations with iterative reasoning.
    \item \textit{DB-Iterative}: A relational-only variant with iterative reasoning.
    \item \textit{DB-OneShot}: A relational-only variant using one-shot reasoning.
    \revision{\item \textit{Graph-Iterative}: A graph-only variant with iterative reasoning.}
    \revision{\item \textit{Graph-OneShot}: A graph-only variant using one-shot reasoning.}
\end{itemize}

The one-shot variants still require two LLM calls (initial query generation and final answer synthesis) but do not enter the iterative loop. \revision{The graph-only variants serve as reference configurations that reflect previous LLM-based IFC querying approaches, which employ either one-shot \citep{iranmanesh2025llm, pacheco2024bimconverse,pan2026llm} or iterative reasoning \citep{Avgoren2025enhancing}. The relational-only variants serve as an additional single-representation reference, isolating the effect of representation choice independently of any specific prior system. Rather than directly comparing against the original systems, we reconstruct their defining design choices within our pipeline, as a like-for-like external comparison is not feasible. The reviewed systems do not release their source code or evaluation query sets, and their underlying schemas also vary. Reconstructing the single-representation and reasoning configurations within our pipeline enables all variants to be evaluated using the same IFC model, LLM, preprocessing procedures, and query set. This ensures that any differences observed between variants are attributable to the choice of data representation and reasoning strategy, rather than to differences in the underlying data.} The ablation study therefore examines the impact of incorporating the hybrid representation and iterative reasoning under consistent data and experimental conditions.

\section{Tests and Evaluation}
\label{sec:tests-evaluation}
This section first reports the generation times for converting IFC models into databases, followed by the results of the query evaluation and ablation study.

\subsection{IFC-to-Database Generation Time}
\label{sec:generation-times}

We first report the cost of the preprocessing step on which all subsequent queries depend: transforming an IFC model into its relational and graph representations. As the number of building elements increases, the computational cost of generating the databases can grow significantly. To provide a baseline, we evaluate 10 IFC models of varying sizes and complexities, measuring the time required to generate both databases on an Apple M1 Pro chip\footnote{8-core CPU, 14-core GPU and 16-core Neural Engine}. Results are summarized in Table \ref{tab:generation-times}.

Results show that generating the relational database is the most time-consuming step, as it involves parsing building elements, extracting their geometries, normalizing units, calculating centroids and bounding boxes, storing property sets, and inserting detailed geometric data. As the model size increases, the database generation time also increases, confirming the overhead associated with transforming large IFC models. For example, relational database generation required 2.5526 seconds for the smallest model (108 building elements; 13 navigable elements), 5.8264 seconds for the mid-sized model (1,015 building elements; 218 navigable elements), and 219.4374 seconds for the largest (25,053 building elements; 2,580 navigable elements). In contrast, graph generation primarily involves adjacency analysis and remains significantly faster, taking only 0.0005 seconds, 0.0713 seconds, and 3.98 seconds for the smallest, mid-sized, and largest models, respectively. In terms of time complexity, graph generation in our implementation has a complexity of $O(n^2)$, where $n$ is the number of navigable elements. For reference, the selectively scoped approach adopted in this work generates the relational and graph representations for Building B1 ("FZK Haus") in 2.56 and 0.0005 seconds respectively. This is not a like-for-like comparison with full-schema pipelines, because the extracted data and the hardware differ; \citet{zhu2023ifc} report 1,798 seconds to represent the same model as a single full-schema graph using IFC-GRAPH.

\begin{landscape}
\begin{table}
\centering
\footnotesize
\caption{\revision{Time (in seconds) required to transform IFC models of varying sizes into relational (RDB) and graph databases (GDB). Models are classified as smallest, mid-sized, or largest based on their number of navigable elements (NE). ``IFC File Lines'' refers to the number of lines in the raw IFC (STEP) file; ``BE'' and ``NE'' denote building elements and navigable elements, respectively. $^\dagger$includes both IfcStair and IfcStairFlight.}}
\label{tab:generation-times}
\begin{tabular}{ccc|lllllll|cc}
\toprule
\textbf{IFC File Lines} & \textbf{BE} & \textbf{NE} & \textbf{Rooms} & \textbf{Doors} & \textbf{Stairs$^\dagger$} & \textbf{Windows} & \textbf{Slabs} & \textbf{Beams} & \textbf{Storey} & \textbf{RDB} & \textbf{GDB} \\
\midrule
82.3k & 108 (\textit{smallest})    & 13                          & 7     & 5     & 1      & 11      & 4     & 51    & 2      & 2.5526       & 0.0005       \\
38.9k & 184                        & 49                          & 21    & 22    & 6      & 30      & 6     & 17    & 4      & 5.1399       & 0.0026       \\
41.6k & 168                        & 37                          & 19    & 16    & 2      & 16      & 8     & 13    & 4      & 3.4666       & 0.0016       \\
49.6k & 647                        & 163                         & 82    & 77    & 4      & 206     & 26    & 0     & 5      & 9.6906       & 0.0152       \\
1199k & 25,053 (\textit{largest})  & 2,580                       & 1103  & 1271  & 203    & 1725    & 99    & 0     & 11     & 219.4374     & 3.9800       \\
38.9k & 223                        & 39                          & 21    & 14    & 4      & 24      & 21    & 8     & 4      & 1.5339       & 0.0022       \\
62.9k & 1,345                      & 207                         & 99    & 102   & 6      & 69      & 2     & 0     & 3      & 4.6462       & 0.0398       \\
529k & 180                        & 41                          & 14    & 21    & 6      & 17      & 13    & 0     & 6      & 1.5733       & 0.0036       \\
383k & 2,069                      & 279                         & 139   & 122   & 18     & 68      & 60    & 512   & 5      & 11.6476      & 0.1193       \\
2487k & 1,015 (\textit{mid-sized}) & 218                         & 105   & 105   & 8      & 128     & 6     & 0     & 6      & 5.8264       & 0.0713\\
\bottomrule
\end{tabular}
\end{table}
\end{landscape}

\subsection{Query Evaluation}
\label{sec:experiments}

\subsubsection{Overall Results}
\revision{Results from the evaluation on Buildings B1 and B2 are summarized in Table \ref{tab:results}. Among the evaluated LLMs, GPT-OSS 120B outperformed the other LLMs in terms of correctly providing the final answers to the questions in the evaluation query set. It achieved an accuracy of 96\%, followed by its compact version, GPT-OSS 20B, which achieved an accuracy of 93\%. Gemma 4 31B and Llama 3.3 70B achieved accuracies of 87\% and 82\%, respectively. The proprietary baseline considered in this work, GPT-5.6 Luna, achieved an accuracy of 90\%. Table \ref{tab:results} further reports, for each LLM, token usage, the number of LLM calls, and the number of SQL/Cypher queries for both query types.}

\revision{All evaluated LLMs achieved relatively strong performance on SQL-based queries, with per-building accuracies generally close to 90\%. In contrast, graph-based performance varied across models. GPT-OSS 120B and GPT-5.6 Luna maintained comparatively high graph querying accuracy (>85\%), whereas Gemma 4 31B and Llama 3.3 70B achieved 57\%, and GPT-OSS 20B achieved 71\% on Building B2. This suggests that generating correct Cypher queries over the IFC-derived graph is more challenging than generating SQL for attribute-centric queries.}

\begin{landscape}
\begin{table}[t!]
\footnotesize
\caption{\revision{Performance of IfcLLM on Buildings B1 and B2. ``Acc.$_{O}$'' is overall accuracy and ``Acc.$_{Q}$'' is accuracy across a query type.}}
\label{tab:results}
\begin{tabular}{llllllllll}
\toprule
\textbf{LLM} & \textbf{Acc.$_{O}$}           & \textbf{Bldg.} & \textbf{Query Type} & \textbf{Acc.$_{Q}$} & \textbf{Tokens$_{Inp}$}          & \textbf{Tokens$_{Out}$}          & \textbf{LLM Calls}      & \textbf{\# of SQL}           & \textbf{\# of Cypher}        \\ \midrule
\multicolumn{9}{l}{\textit{\underline{Open-weight LLMs}}}\\
\multirow{4}{*}{\textbf{GPT-OSS 120B}} & \multirow{4}{*}{0.96}  & \multirow{2}{*}{B1}       & SQL     & 1        & 3k      & 834     & 3.70   & 1.87   & -- \\
 &  &        & GRAPH   & 1        & 4k     & 1k    & 4.14   & 0.29   & 2.00   \\ \cdashline{3-10}
 &  & \multirow{2}{*}{B2}       & SQL     & 0.91     & 4.1k     & 1.2k    & 3.87   & 2.48   & -- \\
 &  &        & GRAPH   & 1        & 4.6k   & 1k    & 4.43   & 1.00   & 1.86   \\ \midrule
\multirow{4}{*}{\textbf{GPT-OSS 20B}} & \multirow{4}{*}{0.93}   & \multirow{2}{*}{B1}       & SQL     & 1        & 2.8k    & 1.4k   & 3.57   & 1.65   & -- \\
 &   &        & GRAPH   & 0.86     & 3.2k   & 811     & 3.57   & -- & 1.57   \\ \cdashline{3-10}
 &   & \multirow{2}{*}{B2}       & SQL     & 0.96     & 8.2k   & 1.7k   & 3.83   & 1.96   & -- \\
 &   &        & GRAPH   & 0.71     & 3.4k    & 849     & 3.43   & -- & 1.43   \\ \midrule
\multirow{4}{*}{\textbf{Gemma 4 31B}} &\multirow{4}{*}{0.87}   & \multirow{2}{*}{B1}       & SQL     & 1        & 2.5k    & 99       & 3.09   & 1.13   & -- \\
 &   &        & GRAPH   & 0.57     & 2.6k    & 89       & 3.14   & 0.29   & 1.00   \\ \cdashline{3-10}
 &   & \multirow{2}{*}{B2}       & SQL     & 0.91     & 3.8k   & 174      & 3.09   & 1.13   & -- \\
 &   &        & GRAPH   & 0.57     & 3.2k    & 121      & 3.29   & 0.29   & 1.14   \\ \midrule
\multirow{4}{*}{\textbf{Llama 3.3 70B}} & \multirow{4}{*}{0.82} & \multirow{2}{*}{B1}       & SQL     & 0.91     & 2.3k    & 362      & 2.87   & 1.30   & -- \\
 & &        & GRAPH   & 0.57     & 2.4k    & 429     & 3.00   & 0.29   & 1.00   \\ \cdashline{3-10}
 & & \multirow{2}{*}{B2}       & SQL     & 0.87     & 3.1k    & 416     & 3.00   & 1.61   & -- \\
 & &        & GRAPH   & 0.57     & 3.2k    & 539     & 3.00   & 0.29   & 1.57   \\ \midrule
 \multicolumn{9}{l}{\textit{\underline{Proprietary baseline}}}\\
\multirow{4}{*}{\textbf{GPT-5.6 Luna}} & \multirow{4}{*}{0.90}  & \multirow{2}{*}{B1}       & SQL     & 0.91     & 3.2k     & 317     & 3.52   & 1.52   & -- \\
 &  &        & GRAPH   & 0.86     & 2.4k    & 225     & 3.14   & -- & 1.14   \\ \cdashline{3-10}
 &  & \multirow{2}{*}{B2}       & SQL     & 0.87     & 44k     & 641     & 4.09   & 2.13   & -- \\
 &  &        & GRAPH   & 1        & 2.9k    & 235      & 3.29   & -- & 1.29   \\ \bottomrule
\end{tabular}
\end{table}
\end{landscape}

\revision{The LLMs also differed in token usage. Gemma 4 31B was the most economical, generating $\approx$90--170 output tokens per query. The GPT-OSS models produced longer responses while achieving higher accuracy. The number of LLM calls was broadly similar across LLMs, at $\approx$3--4 calls per query. This indicates that the differences in token consumption were primarily from prompt and response length rather than from a substantially different number of reasoning steps. Token usage was generally stable for graph queries but in some cases it increased sharply for SQL queries involving large result sets. For example, on Building B2, queries requesting property information for walls, U-values, or door materials produced large tabular results that were passed back to the LLM, resulting in $\approx$44k input tokens per SQL query for GPT-5.6 Luna and $\approx$8k for GPT-OSS 20B. Because these increases in input tokens were associated with particular queries rather than consistently with a specific model, they indicate that result size is an important determinant of inference cost. In deployment, these scenarios could be mitigated by either truncating or summarizing the query results.}

\revision{Among the LLMs evaluated, GPT-OSS 120B
yields the top-performing IfcLLM configuration. The following section examines this configuration in more detail. We first summarize its retrieval behaviour in Buildings B1 and B2, then extend its evaluation to Building B3, to assess whether the same retrieval behaviour generalizes to a larger and structurally more complex IFC model.}

\subsubsection{Top-performing IfcLLM Configuration}

\textit{Buildings B1 and B2.} With GPT-OSS 120B, the system correctly answered all 30 queries for Building B1 and 28 queries for Building B2. The query ``\textit{What is the total building height?}'' required a new run (without modification) to produce the correct response. For the query ``What are all the properties defined for walls?'', the system encountered a \textit{context length exceeded} error due to an SQL query that returned all walls and their properties. To illustrate the system's behaviour, representative interactions for Building B1 are provided in Table~\ref{tab:sample-responses}.

\begin{table}
    \centering
    \footnotesize
    \caption{Sample interactions with IfcLLM on Building B1 (FZK Haus; Figure \ref{fig:hauz-ifc}).}
    \label{tab:sample-responses}
    \begin{tabular}{p{4cm}|p{9.5cm}}
    \toprule
    \textbf{User Query} & \textbf{IfcLLM Response}\\
    \midrule
    How many rooms are on each storey?     &  Sure! Based on the data you provided, the number of rooms on each storey is:\newline
    - Erdgeschoss (ground floor) – 6 rooms\newline
    - Dachgeschoss (top floor/attic) – 1 room \\
    \midrule
    Which rooms are located on the top-most storey?  & The only room that appears on the highest storey is room 7, which is located on the storey named Dachgeschoss. \\
    \midrule
    Which walls are external and internal ones, list them by floors? & Walls grouped by floor and type:\newline
    [..categorized list of walls..]\newline
    Explanation:\newline
The query returned each wall together with the property IsExternal. When IsExternal = True the wall is classified as External; otherwise it is Internal. Duplicate rows (the first query) were removed in the second query, giving a clean list of unique walls per storey.\newline
So, on the Erdgeschoss you have four external walls and five internal walls, while the Dachgeschoss contains only the four external walls shown above. \\
\midrule
Which rooms are connected by doors to Room ``1"? & Room 1 is directly connected by doors to the following rooms: Room 2, Room 3, and Room 4. These are the only rooms that have a door linking them to room 1 according to the query results.\\
\midrule
What is the shortest path between Room ``6" and Room ``7"? & The query shows that the minimal-hop path consists of three connections (three “hops”). The nodes you travel through are:\newline
- Room 6 (starting point)\newline
- Room 5\newline
- Wendeltreppe – the stairwell\newline
- Room 7 (destination)\newline
Total distance (sum of the three segments) $\approx$ 10.32 units.\newline
Thus, the shortest path between room 6 and room 7 goes through room 5 and the Wendeltreppe stair, requiring three hops and covering roughly 10.3 distance units. \\ \midrule
 Which rooms are isolated (not reachable from any other room)? & There are no isolated rooms in the graph – every room can be reached (directly or indirectly) from at least one other room.\\
    \bottomrule
    \end{tabular}
\end{table}

\begin{figure}
    \centering
    \begin{subfigure}[b]{1\linewidth}
        \centering
        \includegraphics[width=\linewidth]{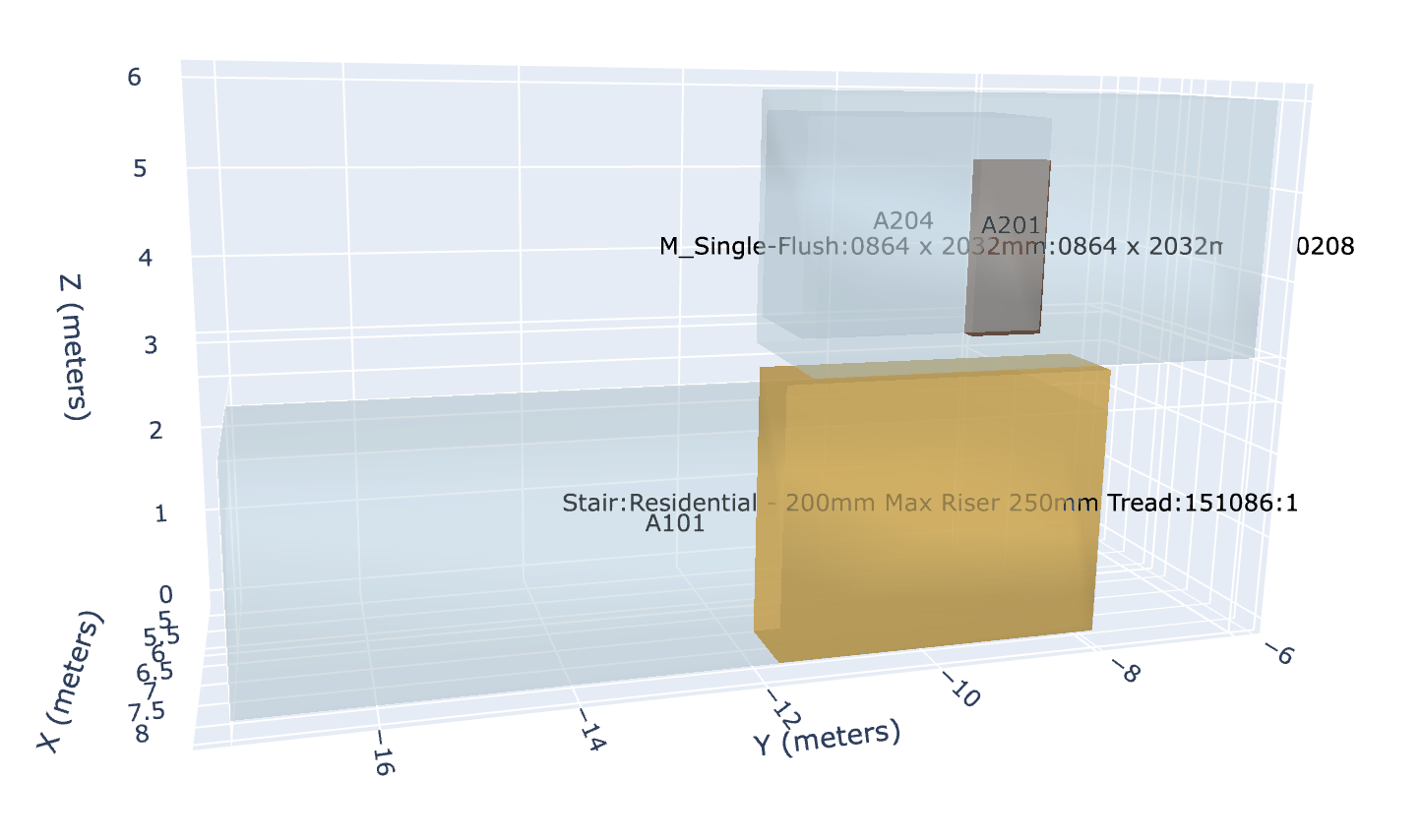}
        \caption{Bounding box representation.}
        \label{fig:selected_elements}
    \end{subfigure}
    \hfill
    \begin{subfigure}[b]{1\linewidth}
        \centering
        \includegraphics[width=\linewidth]{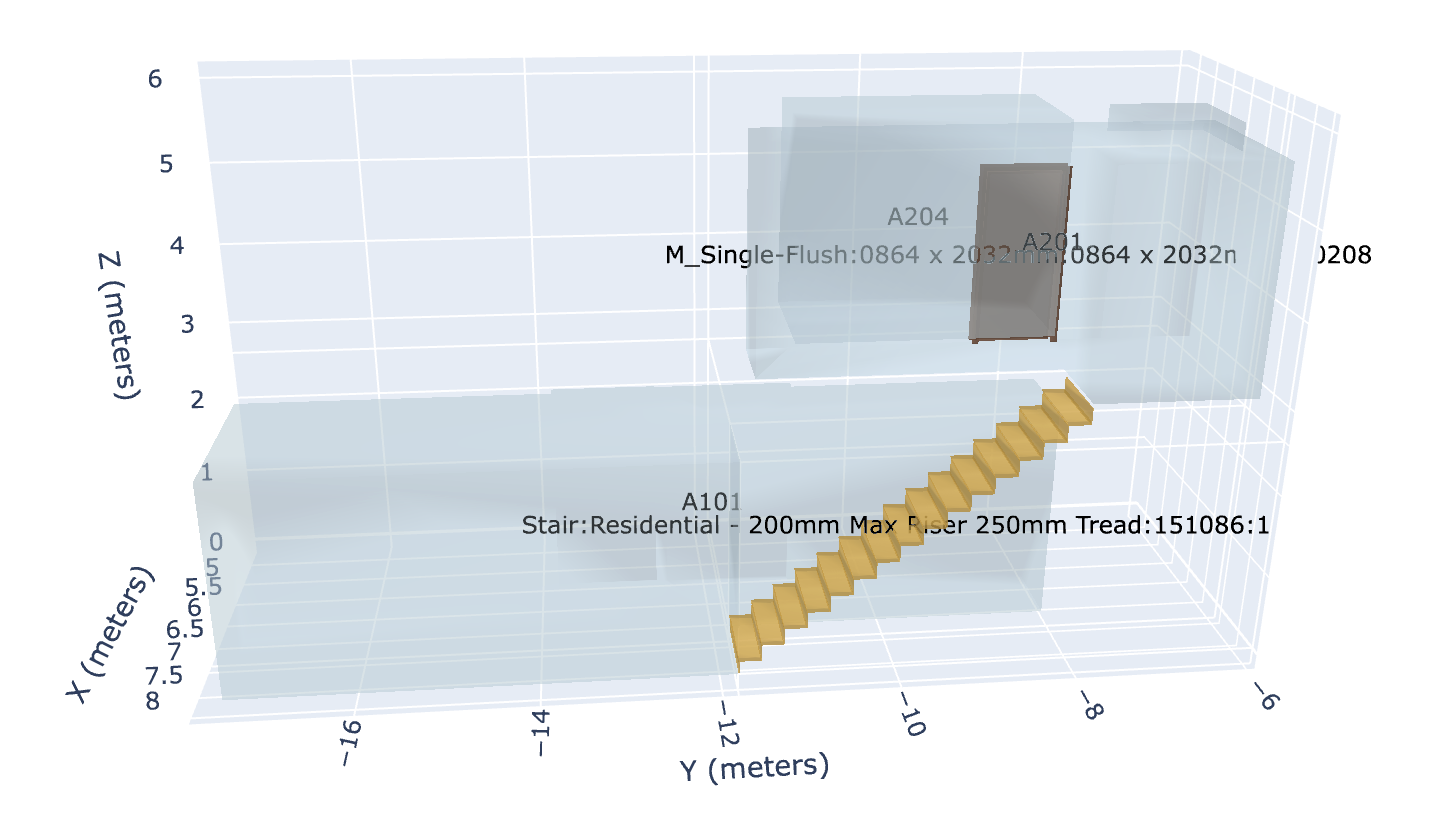}
        \caption{Full geometric representation.}
        \label{fig:selected_elements_real_geom}
    \end{subfigure}
    \caption{Visualization of the IfcLLM-generated shortest path between Room A101 and Room A204 in Building B2, shown with both bounding boxes and full geometry.}
    \label{fig:shortest_path}
\end{figure}

Overall, the system successfully handled the curated set of queries, demonstrating functional capability in semantic, hierarchical, and spatial reasoning within the defined evaluation scope. It correctly computed building-level metrics such as total height using storey elevations and interpreted spatial references such as \textit{top-most} and \textit{bottom-most}. Numerical outputs (e.g., room volumes) were appropriately formatted, improving readability. The system also processed multilingual data; for example, German floor names in Building B1 (e.g., Erdgeschoss, Dachgeschoss) were correctly interpreted and translated in responses, reflecting the underlying LLM's general language capabilities. This behaviour may be beneficial in practice, where IFC models from different regions often contain mixed-language metadata. The graph representation supported efficient traversal across building elements. Connectivity and navigation queries, including adjacency and pathfinding, were answered correctly. The system identified connected rooms, checked spaces if they are isolated, and computed shortest paths. These results can also be visualized, as shown in Figure~\ref{fig:shortest_path}, using both bounding box and full geometric representations. The system further demonstrated consistent performance in attribute retrieval and aggregation tasks. It correctly counted elements, extracted properties, and grouped results by storey where applicable. It also handled missing or incomplete data appropriately. For instance, when attributes such as fire ratings were absent, the system reported the missing values instead of generating incorrect outputs. In Building B2, missing \textit{name} and \textit{description} fields were identified, and the absence of certain element types (e.g., slabs or ceilings on the lowest floor) was correctly reported.

\textit{Building B3.} The system correctly handled the same set of queries for Building B3, successfully answering 29 out of 30 queries (96.7\% accuracy). The one failed query ``Which doors are exit doors?" caused the system to retrieve all door data rather than filtering for exit-related properties, resulting in a response that exceeded the context limit. Apart from this, the system correctly identified missing properties (e.g., building name and description), computed building height, and retrieved element counts both globally and per storey. It also correctly interpreted spatial references such as \textit{ground}, \textit{bottom-most}, and \textit{second top-most} floors. Graph-based queries were handled consistently, including identifying connected rooms, determining shortest paths, and detecting disconnected spaces. The system also handled incomplete data and schema inconsistencies, such as the absence of a roof element or missing fire-rating attributes. Likewise, it identified missing or incorrectly populated attribute values; for example, the absence of a numeric or class-based fire rating for doors.

Overall, the results demonstrate that the hybrid representation enables the system to support both attribute-centric and spatial queries across the evaluated IFC models, without requiring direct ingestion of raw IFC texts by the LLM.

\subsection{Ablation Study}
\label{ablation}

Next, we perform an ablation study that isolates the impact of the framework's core design choices. All ablation variants (\textit{IfcLLM}, \textit{DB-Iterative}, \textit{DB-OneShot}, \textit{Graph-Iterative}, \textit{Graph-OneShot}; discussed previously in Section \ref{evaluation-criteria}) are evaluated on the same IFC model (Building~B1). They share an identical preprocessing pipeline and use the same top-performing LLM, GPT-OSS 120B.

\textit{For relational-only variants.} Adjacency between navigable building elements (e.g., rooms (spaces), doors, and stairs) is derived from geometry, as per Algorithm~\ref{algo:adjacency_detection}. This information is stored in two relational tables: \textit{spatial\_nodes} and \textit{spatial\_edges}. The \textit{spatial\_nodes} table represents building elements with attributes such as a unique identifier and element type, while \textit{spatial\_edges} encodes adjacency using \textit{from\_node}, \textit{to\_node}, \textit{distance}, and a binary \textit{is\_vertical} flag. 

\begin{figure}
    \centering

    \begin{subfigure}{\textwidth}
        \centering
        \hspace*{-3cm}
        \includegraphics[width=0.75\linewidth]{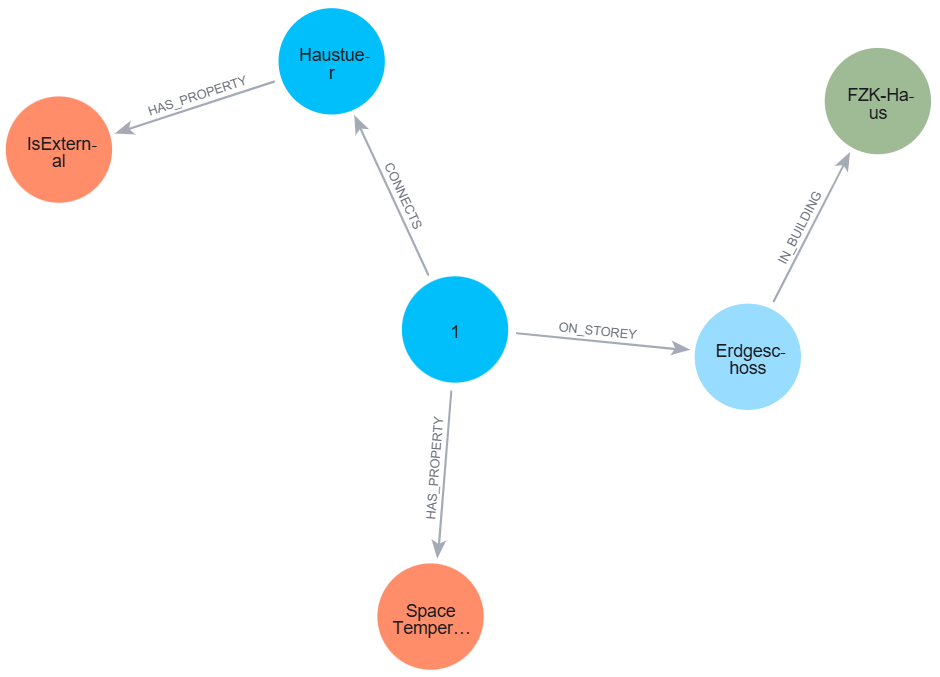}
        \caption{Showing node types: building, storey, room, door, property, and related relationships.}
        \label{fig:graph-only-small}
    \end{subfigure}

    \vspace{0.2cm}

    \begin{subfigure}{\textwidth}
        \centering
        \hspace*{-3cm}
        \includegraphics[width=1.35\linewidth]{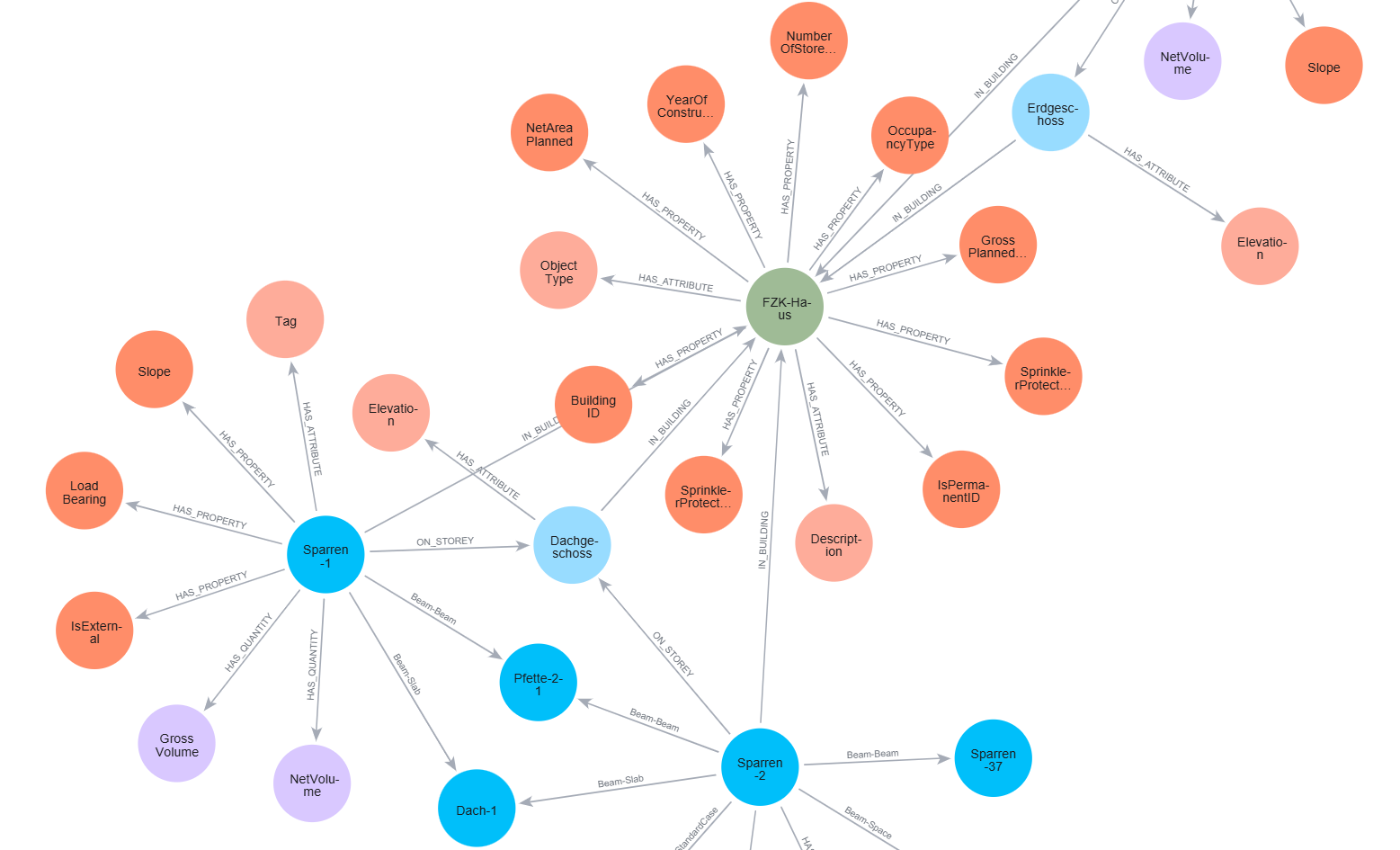}
        \caption{Showing node types: building, storey, attribute, property, quantity, beam, and related relationships.}
        \label{fig:graph-only-large}
    \end{subfigure}

    \caption{\revision{Representative subgraphs of the IFC-derived property graph for Building B1.}}
    \label{fig:graph-only}
\end{figure}

\revision{\textit{For graph-only variants.} The IFC model is transformed directly into a graph in which every building element (IfcElement) and space (IfcSpace) becomes a node, together with dedicated \textit{Building} and \textit{Storey} nodes. Each node carries the same information as its relational counterpart. The attributes, property sets, and quantity sets are attached as separate \textit{Attribute}, \textit{Property}, and \textit{Quantity} nodes, and the spatial containment hierarchy is encoded through ON\_STOREY and IN\_BUILDING relationships. Elements and spaces whose triangulated meshes clash are linked, capturing physical connectivity directly from geometry. Separately, adjacency between navigable elements (rooms, doors, and stairs) is derived using the same procedure as the other variants (Algorithm~\ref{algo:adjacency_detection}) and is stored as CONNECTS relationships, each annotated with a distance and a binary vertical-connection flag, so that the spatial relationships remain consistent with the relational representation. This broader graph is required so that the graph-only variants can answer both SQL-based and graph-based queries present in the evaluation query set. Figure \ref{fig:graph-only} shows representative subgraphs of the IFC-derived property graph for Building B1.}

\subsubsection{Ablation Results}
Table~\ref{tab:ablation} summarizes the results on the evaluation query set. \textit{IfcLLM} and the relational-only variant with iterative reasoning (\textit{DB-Iterative}) achieved 100\% accuracy. The remaining three variants, \textit{DB-OneShot}, \textit{Graph-OneShot} and \textit{Graph-Iterative}, reached 93\%. \revision{Because the graph-only variants reconstruct the single-representation design of prior LLM-based IFC querying systems under identical data, LLM, and preprocessing conditions (Section~\ref{evaluation-criteria}), these results also serve as a controlled comparison of IfcLLM against those approaches.}

\revision{Performance on SQL-based queries was high for all variants with access to the relational schema. \textit{IfcLLM}, \textit{DB-Iterative}, and \textit{DB-OneShot} each achieved 100\% accuracy queries, whereas the graph-only variants achieved 96\%. The errors in the graph-only variants occurred on aggregation queries that required Cypher. For graph-based queries, \textit{IfcLLM} and \textit{DB-Iterative} again achieved 100\% accuracy, while \textit{Graph-OneShot} and \textit{Graph-Iterative} achieved 86\%, and \textit{DB-OneShot} achieved 71\%. \textit{DB-OneShot} failed on two spatial queries involving distance computation and navigability. Without iterative reasoning, it could not reformulate the erroneous queries, whereas \textit{DB-Iterative} successfully reformulated them and produced the correct results. The errors in the two graph-only variants were different: \textit{Graph-OneShot} failed on a shortest-path query, which could not be corrected without iterative reasoning, while \textit{Graph-Iterative} failed to provide the correct count for a query asking for the rooms with the most doors. These errors are discussed further during error analysis (Section \ref{sec:error-analysis}).}

\begin{figure}
    \centering
    \begin{subfigure}[t]{\textwidth}
        \centering
        \includegraphics[width=0.9\linewidth]{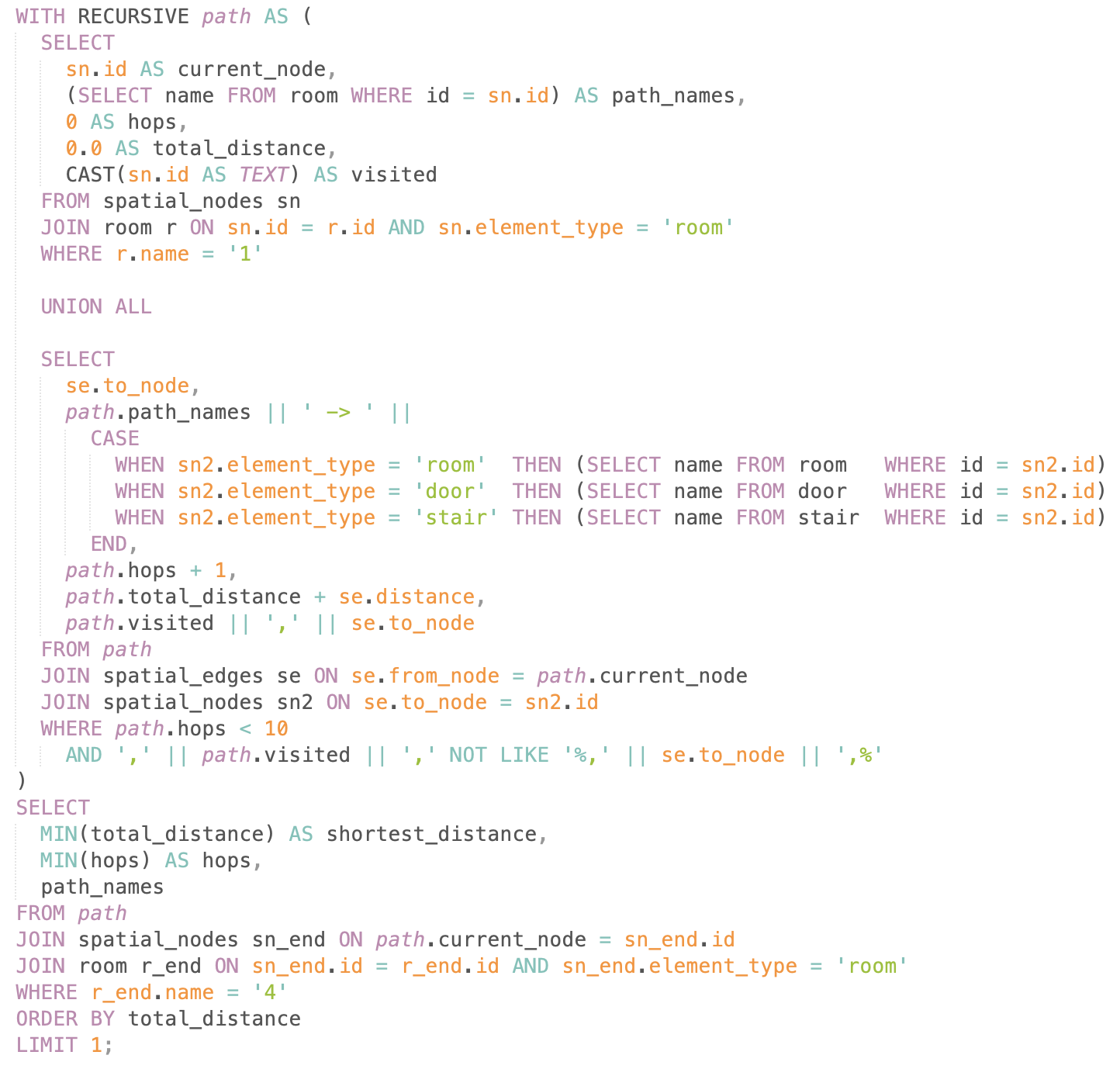}
        \caption{SQL Query}
        \label{fig:sql-example}
    \end{subfigure}

    \vspace{2em}

    \hspace*{-0.05\textwidth}
    \begin{subfigure}[t]{\textwidth}
        \centering
        \includegraphics[width=1.1\linewidth]{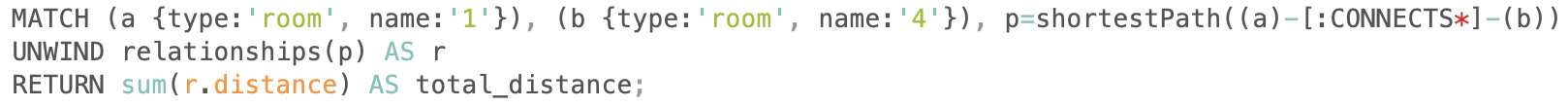}
        \caption{Cypher Query}
        \label{fig:cypher-example}
    \end{subfigure}
    \caption{Comparison of equivalent shortest-path computation queries generated by GPT-OSS 120B to answer ``\textit{What is the distance between room 1 and room 4?}`` for Building B1.}
    \label{fig:sql-vs-cypher}
\end{figure}

\revision{Although \textit{DB-Iterative} matched \textit{IfcLLM} in accuracy, graph-based queries required longer recursive SQL and reformulations, resulting in higher token consumption. On average, for graph-based queries, \textit{DB-Iterative} used $\approx$6k input tokens and $\approx$2.4k output tokens per query, compared with $\approx$4.0k and $\approx$1.0k for \textit{IfcLLM}, respectively. Its system prompt was also longer, at $\approx$3.2k tokens compared with $\approx$1.7k for \textit{IfcLLM}. Figure~\ref{fig:sql-vs-cypher} illustrates this difference using representative SQL and Cypher queries for the same shortest-path computation task. \textit{Graph-OneShot} was the most token-efficient variant, using $\approx$2.4k input and $\approx$500 output tokens per query, but achieved only 93\% accuracy. \textit{Graph-Iterative} did not improve accuracy, while answering SQL-based queries became more expensive, with $\approx$5.6k input and $\approx$1.1k output tokens per query and an average of 4.5 LLM calls.}

\begin{landscape}
\begin{table}
\footnotesize
\caption{\revision{Ablation study results. ``Acc.$_{O}$'' is overall accuracy and ``Acc.$_{Q}$'' is accuracy across a query type.}}
\label{tab:ablation}
\begin{tabular}{lllllllll}
\toprule
\textbf{Variant}         & \textbf{Acc.$_{O}$}               & \textbf{Query Type} & \textbf{Acc.$_{Q}$} & \textbf{Tokens$_{Inp}$}      & \textbf{Tokens$_{Out}$}     & \textbf{LLM Calls}   & \textbf{\# of SQL}       & \textbf{\# of Cypher}    \\
\midrule
IfcLLM          & \multirow{2}{*}{1}    & SQL     & 1        & 3k    & 834.26     & 3.70   & 1.87   & -           \\
                &                       & Graph   & 1        & 4k   & 1k    & 4.14   & 0.29   & 2.00   \\ \midrule
DB-OneShot      & \multirow{2}{*}{0.93} & SQL     & 1        & 3.5k    & 390.30      & 2.00   & 1.00   & -           \\
                &                       & Graph   & 0.71     & 3.5k     & 840.57     & 2.00   & 1.00   & -           \\ \midrule
DB-Iterative    & \multirow{2}{*}{1}    & SQL     & 1        & 4.3k    & 757.13     & 3.57   & 1.70   & -           \\
                &                       & Graph   & 1        & 6k   & 2.4k   & 4.86   & 3.29   & -           \\ \midrule
Graph-OneShot   & \multirow{2}{*}{0.93}  & SQL     & 0.96     & 2.3k     & 537.48     & 2.00   & -           & 1.00   \\
                &                       & Graph   & 0.86     & 2.4k    & 553.57     & 2.00   & -           & 1.00   \\ \midrule
Graph-Iterative & \multirow{2}{*}{0.93}  & SQL     & 0.96     & 5.6k   & 1.1k    & 4.52   & -           & 2.91   \\
                &                       & Graph   & 0.86     & 3.4k   & 633.29     & 3.29   & -           & 1.29   \\ \bottomrule
\end{tabular}
\end{table}
\end{landscape}

\section{Discussion}
\label{sec:discussion}
This section provides an analysis of the evaluation results, including observed errors, implications, limitations, and directions for future research.

\subsection{Error Analysis}
\label{sec:error-analysis}
\revision{We performed an error analysis on the query logs from all evaluations including the ablations to understand \textit{why} the system fails. We observed most errors to be \textit{recoverable}: the SQL/Cypher query returns an error or an empty result, and the retry loop fixes it on a later attempt. A few are \textit{not recoverable}: they either stop the reasoning loop (results too large for the context window) or return a wrong but plausible answer with no error to trigger a retry. The errors can be categorized into six groups, we discuss each with representative examples.}

\subsubsection{Output formatting errors} 
\revision{The most frequent and least severe failure involved syntactically valid queries being generated in an invalid format. LLMs were explicitly instructed to follow a prescribed output format (e.g., for SQL queries: \lstinline|SQL_NEEDED: <SELECT ...>|). The workflow then processes the routing token (i.e., SQL\_NEEDED) and the query (SELECT ...). However, the most common failure was enclosing the query in Markdown code fences (e.g., \lstinline|```sql SELECT ... ```|) rather than returning the bare query required by the workflow. Another failure involved including routing token or explanatory text within the query. For example, for the query \emph{``Which rooms are isolated?''}, Llama~3.3~70B embedded explanatory text within its Cypher query (e.g., This query will return ... connected to them: MATCH ...), which was eventually rejected by the workflow because it violated the required output format. Similarly, GPT-OSS models occasionally prefixed queries with routing tokens. These failures reflect instruction-following rather than reasoning limitations, and the iterative loop corrected the large majority of them on the subsequent attempt.}

\subsubsection{Schema and query dialect mismatch} 
\revision{These failures arise when an LLM assumes a schema structure or query dialect that is inconsistent with the backend. We observed hallucinated relations, such as a non-existent \texttt{property\_value} table joined through an invented \texttt{property\_value\_id} foreign key, as well as hallucinated columns, such as \texttt{door.fire\_rating}, whereas fire rating is actually stored as a row in the \texttt{property} table. Despite being explicitly instructed that the relational database uses SQLite, LLMs also occasionally generated queries using PostgreSQL-specific syntax. For example, the case-insensitive \texttt{ILIKE} operator resulted in a \texttt{near "ILIKE": syntax error} when executed against SQLite. These errors were generally recoverable because the database engine provided explicit error messages that the LLM could use to refine its query. For instance, in one case, an LLM assumed that every element table exposed a particular column; after receiving a \texttt{no such column} error, it inspected the schema through \texttt{sqlite\_master} and reissued a corrected query.}

\subsubsection{Unsuitable backend} 
\revision{The single-representation ablation variants expose a category of error that the hybrid design largely avoids: issuing a query to a representation that is less suited to it. When the relational-only variant was posed connectivity and navigation queries, it repeatedly attempted recursive path computations that were rejected by the workflow; 65.2\% of the SQL statements it generated for graph-based queries were invalid. Conversely, the graph-only variant required repeated reformulations, with 64.2\% of the Cypher it generated for SQL-based queries being invalid. The hybrid system, by contrast, routed each query class to its intended backend: it generated no Cypher for SQL-based queries and issued SQL for graph-based queries only twice. Measured on the backend it actually used, 28.6\% of its Cypher for graph-based queries and 7.0\% of its SQL for SQL-based were invalid. These observations support the routing rationale of this work: attribute retrieval, aggregation, and hierarchy queries are more reliably served by the relational layer, and spatial traversal, adjacency, and shortest-path queries by the graph layer. Allowing the LLM to select the representation eliminates a failure mode present in single-representation systems.}

\subsubsection{Result-size overflow}
\label{sec:result-overflow}
\revision{The most consequential errors arose from queries that were individually valid but retrieved substantially more data than required by the question. Requests such as ``\textit{What are the materials or fire ratings of all doors?}'' and ``\textit{What is the U-value of external walls?}'' led some LLMs to retrieve all properties from the \texttt{property} table, producing large tabular results that were subsequently fed back into the prompt. This caused input token counts to increase to $\approx$139k for a single door-materials query on Building~B2 and up to $\approx$428.7k for a wall-property query with GPT-5.6~Luna. For GPT-OSS 120B, when the result exceeded the model's context window, the reasoning loop terminated before an answer could be produced, as observed for the wall-properties query on Building~B2 and the exit-doors query on Building~B3. Unlike the previous categories, these failures are not self-correcting, as a retry typically reissues the same broad query and produces another oversized result.}

\subsubsection{Semantic errors}
\label{sec:semantic-errors}
\revision{The most difficult errors to detect are syntactically valid queries that execute successfully but perform incorrect data retrieval. A clear example is the total-building-height query, for which an LLM computed \texttt{MAX(elevation)} over the storey table (or, in another case, summed the storey elevations) instead of computing the difference between the highest and lowest elevations. On Building~B1, this produced a correct value because the ground storey was located at $z\approx0$, thereby masking the error; on Building~B2, this issue was relevant as the lowest elevation is below 0. A related data-level error is the incorrect room-to-room connection caused by overlapping bounding boxes of irregularly shaped rooms (discussed further in Section~\ref{sec:enhancing-backend}). Because such errors are silent, they provide the motivation for incorporating an explicit answer-validation stage.}

\subsubsection{Errors and accuracy} 
\revision{A notable observation is that a high number of invalid queries does not necessarily imply lower accuracy. GPT-OSS~120B generated the highest number of invalid queries in the hybrid setting (35 invalid SQL and 9 invalid Cypher across Buildings~B1 and B2), yet achieved the highest overall accuracy by effectively leveraging the retry-and-refine loop. In contrast, Gemma~4~31B produced relatively few invalid queries (2 invalid SQL and 1 invalid Cypher query) but frequently returned answers without further iteration, causing its error to manifest as incorrect final answers rather than detectable execution failures. This observation suggests that an LLM that produces detectable errors and can recover through iterative refinement may be preferable to one that rarely generates invalid queries but lacks the ability to detect and correct its own mistakes.}

\subsection{Implications, Limitations and Future Directions}
\label{sec:implications}
The evaluation results have several implications for LLM-based IFC querying, which are discussed below along with the limitations and directions for future research.

\subsubsection{The Hybrid Representation}
\revision{Within the ablation study on Building B1, the results indicate that the performance advantage of the hybrid representation arises from matching the representation to the query type. The relational-only variant (without iterative reasoning) performed reliably for SQL-based queries but was less effective for graph-based queries. Enabling iterative reasoning allowed it to achieve comparable accuracy to the hybrid system, but required longer recursive SQL queries and reformulations, resulting in higher token consumption. Conversely, the graph-only variant was well suited to graph-based queries, while for SQl-based queries its token consumption was highest among the evaluated variants.}

\revision{The hybrid design avoids these costs by routing queries to the representation best suited to their requirements rather than relying solely on iterative reasoning to compensate for the limitations of a single representation. As noted during error analysis, this is reflected in IfcLLM issuing only two SQL queries for graph-based queries, no Cypher for SQL-based queries, and a lower rate of invalid queries on the backend actually used for each query type, while matching the accuracy of the iteration-enabled relational-only variant with substantially lower token consumption.}

\revision{These findings address \textbf{RQ1} and \textbf{RQ2} directly. For RQ1, matching each query to its better-suited backend makes the hybrid the most efficient among the variants that reached 100\% accuracy. For RQ2, the iterative-versus-one-shot comparison shows that retry-and-refine reasoning is what improves the performance of the relational-only variant to the same accuracy as the hybrid system, but at a higher token cost. On the graph-only variants, iterative reasoning did not raise overall accuracy (both 93\%): \textit{Graph-OneShot} failed on a shortest-path query, while \textit{Graph-Iterative} failed on an aggregation query that the retry loop did not correct.}

\subsubsection{Open-Weight LLMs}
\revision{Within the scope of the evaluated query set, the overall results suggest that local deployment with an open-weight model does not necessarily require sacrificing query accuracy relative to the proprietary baseline considered in this work. GPT-OSS 120B and GPT-OSS 20B both achieved higher overall accuracy than the proprietary GPT-5.6 Luna baseline considered in this work. Moreover, the gap between the two GPT-OSS variants was relatively small: GPT-OSS 20B achieved 93\% overall accuracy, only three percentage points below GPT-OSS 120B. This may indicate that competitive performance on this query set does not require the largest available open-weight LLM, which could be a relevant consideration for organizations that must keep IFC data on-premises but have limited access to computational resources. Further evaluation across additional buildings, LLMs, and query scenarios is needed to assess how well this pattern generalizes.}

\revision{These observations address \textbf{RQ3}. The choice of open-weight LLM affected all three dimensions considered. It affected accuracy, with GPT-OSS 120B and 20B matching or exceeding the proprietary baseline while Gemma 4 31B and Llama 3.3 70B trailed; efficiency, with models differing mainly in response verbosity rather than number of reasoning steps; and failure behaviour, where models that produced detectable, recoverable errors (e.g., GPT-OSS 120B) ultimately outperformed those that returned fewer invalid queries but more silent, incorrect answers (e.g., Gemma 4 31B).}

\subsubsection{LLM Uncertainty and Hallucination}
\revision{Like any LLM-based querying system, IfcLLM is subject to the non-deterministic nature of the underlying LLM, which can generate hallucinated or incorrect outputs. This behaviour is reflected in the error categories reported in Section~\ref{sec:error-analysis}, where LLMs occasionally did not follow prescribed output formats, hallucinated non-existent tables, columns, or relationships, assumed an incorrect query dialect, or produced valid queries that silently retrieved unintended data. Because queries are executed against a fixed, schema-constrained backend, hallucinated schema elements usually raise explicit execution errors that the retry-and-refine loop can recover from. This does not, however, address silent semantic errors (Section~\ref{sec:semantic-errors}), which execute successfully but do not reflect the user's intent and remain the most difficult failure mode to detect. IfcLLM should therefore be used as a decision-support tool whose outputs may require verification in high-stakes settings.}

\subsubsection{Enhancing the Backends}
\label{sec:enhancing-backend}
\secondrevision{On the graph side, the current approach to generating the connectivity graph can fail in cases where building elements have curvilinear or irregularly shaped geometries. Such elements' bounding boxes may overlap, leading to inaccurate adjacency computations, as observed in the case of Building B2, where the bounding box of room A201 partially overlaps that of room A202 and produces a false open-room connection in the connectivity graph (Figure \ref{fig:duplex-ifc}(c)). A related concern is that this implementation uses bounding box centroids for distance estimation. For a concave space, the centroid of its bounding box may lie outside the actual space geometry, making distances based on these points less representative of spatial distances relevant to navigation and potentially affecting shortest-path results. Together with inaccurate adjacency caused by bounding box overlap, these approximations may reduce the accuracy and reliability of graph-based queries, even when the LLM generates a valid query against the graph backend. Future implementations should therefore aim to develop graph-generation methods grounded in full geometric data.} \revision{As an initial step in this direction, we developed a pipeline for the ablation study that transforms an IFC model into a property graph of IfcElement and IfcSpace based on whether their triangulated meshes clash (see Figure~\ref{fig:graph-only}). We release the graph-generation pipeline (\url{https://pypi.org/project/ifc2graph/}) to support further development and evaluation by the research community.}

On the relational side, the current backend does not implement spatial extensions. Future work could adopt standardised spatial database extensions such as SpatiaLite or PostGIS, which would enable native spatial SQL functions such as \textit{ST\_Intersects} and \textit{ST\_Distance} to replace the current adjacency and distance computation logic, among other additional benefits of incorporating spatial data types. In addition, integrating semantic 3D city models \citep{lamsal2026cityllm} alongside building models could enable more sophisticated analyses across building and urban-scale contexts.

\subsubsection{Cases for Validation}
\label{sec:cases-for-val}
Also discussed in Section \ref{sec:result-overflow}, the \textit{context length exceeded} errors observed for Buildings B2 and B3 point to a common underlying cause: an SQL query that retrieves more data than the question requires, exceeding the LLM's context limit before reasoning can occur. A related but distinct issue is semantic interpretation error discussed in Section \ref{sec:semantic-errors}. Both cases point to two complementary directions for future research: (i) mechanisms for verifying query results or, at minimum, estimating context size before proceeding with full LLM inference, which could directly address the token-length failures observed in this work; and (ii) multi-model or answer-validation frameworks capable of detecting and correcting semantic interpretation errors, such as those observed in the building-height case.

\subsubsection{Context Management}
The current implementation includes all available properties for building elements (up to a maximum of 100) in the system prompt. This is manageable for smaller models such as Building B1, but larger or more property-rich models can define hundreds of distinct properties (Table \ref{tab:stats}), substantially increasing token count regardless of what a given query actually requires. This is an instance of a more general dynamic context management problem: rather than supplying a fixed, query-independent slice of the schema and property space to the LLM, the system prompt should be constructed adaptively based on what the specific query needs.

One direction for addressing this is to parse the user's query to identify and retrieve only the relevant properties and schema elements for inclusion in the system prompt \citep{lamsal2026query2property}, restricting the prompt to contextually relevant information \citep{zhang2024knowgpt}. This would naturally lower the verbosity in the prompt and the LLM inference cost. A further direction is to construct such relevance filtering using domain-adapted embeddings: prior research shows that domain-adapted contextual embeddings yield more semantically meaningful representations than general-purpose models, including in multilingual settings \citep{reimers2020making,lamsal2024crisistransformers,lamsal2024semantically}. An open research question is whether an IFC schema–aware embedding space could be learned to drive dynamic, query-conditioned prompt construction for both property selection and schema summarisation, an avenue we intend to pursue in future work.

\subsubsection{Towards a Benchmark for LLM-based IFC Querying}
\revision{We note that our evaluation is limited in scope, covering three openly available IFC models and 30 query scenarios; although these span a range of model sizes and common query patterns, the findings should be interpreted within this scope and may not generalize to larger industrial models or a broader range of query scenarios.} This limitation partly reflects a recurring challenge in this research area, namely the lack of openly available, complex IFC models and corresponding evaluation resources, which constrains reproducible benchmarking of LLM-based IFC querying systems. The models used in this study are among the limited set of openly available reference models suitable for this purpose, as more complex, real-world commercial models are rarely released publicly due to confidentiality. \revision{This limitation also constrains direct comparison with prior systems, as their source code and evaluation query sets are generally not publicly available. We therefore evaluate IfcLLM against controlled reconstructions of their single-representation designs (Section~\ref{evaluation-criteria}) rather than their original implementations. The availability of shared IFC models, evaluation query sets, and reference implementations would enable direct, like-for-like comparisons. Benchmarking IfcLLM against representations based on ifcOWL and LBD under identical LLM and data conditions would be another interesting direction for future work.}

The 30 query scenarios developed in this work (\ref{all-queries-motivation}) were designed to be parameterizable and reusable: as discussed in Section \ref{sec:evaluation-queries}, each scenario reflects a recurring query pattern (e.g., property retrieval for a named element, filtering by a condition, aggregation per storey, connectivity between named spaces) that generalizes across different IFC models without modification to the underlying query logic. We suggest that this query set, evaluated here across three models of increasing scale and complexity, can serve as an initial benchmark for future researchers developing or comparing LLM-based IFC querying systems, complementing the limited pool of openly available IFC models. Expanding this set with more diverse and challenging queries, and pairing it with curated ground-truth answers across a wider range of semantic, geometric, and spatial reasoning tasks, would further support reproducible evaluation in this area.

\section{Conclusion}
\label{sec:conclusion}

This paper presented IfcLLM, a framework that uses complementary relational and graph representations to support natural language querying of IFC models. An LLM agent integrates both representations through iterative retry-and-refine reasoning, enabling accessible interaction with IFC data without requiring users to understand the underlying schema.

The framework was evaluated across 30 query scenarios, with an additional ablation study examining its core design choices. Within the evaluation scope of this work, IfcLLM achieved high accuracy across semantic, hierarchical, and spatial queries, with the top-performing configuration reaching 96\% overall accuracy. The ablation results show that iterative reasoning can mitigate the limitations of a single representation, while the hybrid approach achieves comparable accuracy with lower token consumption by routing queries to the representation best suited to the task. The evaluation also found that open-weight models can provide competitive performance: GPT-OSS 120B and 20B achieved 96\% and 93\% accuracy, respectively, compared with 90\% for the proprietary baseline considered in this work. This indicates the potential for open-weight models to support local deployment in settings where IFC data cannot be transmitted to external services.

From an AEC perspective, these results suggest that exposing complementary IFC representations to an LLM has the potential to lower the technical barrier to querying and understanding BIM data. It can also support routine BIM analysis tasks and, in some cases, help surface missing or inconsistent data. The 30 evaluated scenarios, applied consistently across three models of different scales and characteristics, are offered as a starting point for benchmarking future LLM-based IFC querying systems. Overall, combining complementary representations with iterative reasoning provides a promising approach to conversational BIM querying, with the directions outlined in Section~\ref{sec:implications} pointing toward more robust and scalable implementations.

\section*{Acknowledgments}
This work was supported by the Australian Research Council (ARC) Industry Transformation Research Hub (ITRH) for Resilient and Intelligent Infrastructure Systems (RIIS) under Grant: ARC ITRP IH210100048, and was done in collaboration with FrontierSI.

\section*{Declaration of AI and AI-assisted Tools in the manuscript preparation process}
During the preparation of this manuscript, the authors used ChatGPT and Claude to assist with improving grammar. These generative AI tools were not used to develop or generate the original scientific content.

\bibliographystyle{elsarticle-harv}
\bibliography{ref}

\newpage
\appendix

\section{System Prompt Template}
\singlespacing
\label{system-prompt}

\footnotesize
\begin{verbatim}
You are an expert assistant for querying and analyzing IFC building models 
using both SQLite and Neo4j. Generate only queries (no prose) in the required format.

RELATIONAL DATABASE SCHEMA: {%sqlite schema}
GRAPH DATABASE SCHEMA: {%neo4j schema}

AVAILABLE ELEMENT TYPES: {%element_types: SQLite, Neo4j}
PROPERTY NAMES: {%property names}

GUIDELINES:
1) Use SQLite for element hierarchy, properties, and counts.
2) Use Neo4j for spatial navigation, adjacency, shortest paths, and connectivity 
among rooms/doors/stairs.
3) In SQLite: centroid is stored as [x, y, z], and bounding-boxes as [[min_x, min_y, min_z],
[max_x, max_y, max_z]]. Similarly, vertices and faces are JSON arrays.
4) Neo4j has no labels; always match nodes by properties, e.g., (n {type:`room'}).
5) Relationships are CONNECTS with properties distance and type.
6) Treat all names as strings; use case-insensitive matching when necessary.
7) Schema may vary; list sample names if uncertain.
8) Use door properties such as "IsExternal" or "FireExit" to check exits.
9) Slabs may also function as roofs (check predefined_type = ROOF).
10) Responses must be precise, not verbose.

SAMPLE QUERIES:
[%sql queries]
[%cypher queries]

RESPONSE FORMAT:
- For SQLite: SQL_NEEDED: <SELECT ...>
- For Neo4j: CYPHER_NEEDED: <MATCH ...>
- For both: include both prefixes, separated by semicolons.

User Query: {%user input}
\end{verbatim}

\newpage
\section{Intermediate Prompt Template}
\label{intermediate-prompt}

\footnotesize
\begin{verbatim}
You are analyzing query results to determine if you need more information to fully 
answer the user's question.

User Query: {%user input}

Queries executed so far: {%queries + results}

ANALYZE:
1. Can you answer the user's question with the current results?
2. Do you need additional information from the database or graph?
3. If yes, what specific query would help?

RESPOND WITH:
- If you can answer: "ANALYSIS_COMPLETE"
- If you need more SQL data: "MORE_SQL_NEEDED: <your_additional_sql_query>"
- If you need more Cypher data: "MORE_CYPHER_NEEDED: <your_additional_cypher_query>"
- If you need both: "MORE_SQL_NEEDED: <sql_query>; MORE_CYPHER_NEEDED: <cypher_query>"

Remember to be specific about what additional information you need.    
\end{verbatim}

\singlespacing
\newpage

\section{30 User Scenarios Motivating the Evaluation Queries}
\label{all-queries-motivation}

\begin{itemize}
    \item \textbf{Building name/description}: A project manager might query the IFC model for the building’s name or description to verify they have the correct building data before further analysis.
    \item \textbf{Number of storeys}: A facility or campus operations manager might query the IFC model for the number of storeys in a building to estimate cleaning workloads, allocate maintenance staff, or plan security patrol coverage across floors.
    \item \textbf{Storey names and elevations}: A construction manager may need to retrieve the names and elevations of all storeys from the IFC to verify that floor labels and heights can be placed on the facade of the building during construction. 
    \item \textbf{Total building height}: A code compliance officer might query the IFC model for the building’s total height to verify it meets local zoning height restrictions.
    \item \textbf{Counts of rooms/walls/doors/windows}: A quantity surveyor could query the IFC model for the total number of rooms, walls, doors, and windows to estimate material quantities and labour requirements during cost estimation and procurement planning.
    \item \textbf{Rooms per storey}: A facilities manager might use the IFC model to determine how many rooms are on each storey, helping allocate maintenance or cleaning resources floor by floor.
    \item \textbf{Ground and top floor elevation}: An engineer or surveyor might query the IFC to retrieve the elevation of the ground floor and the top-most floor to confirm foundation level and roof height with site measurements.
    \item \textbf{Rooms on top-most storey}: A maintenance scheduler or safety coordinator might query the IFC to list which rooms are on the top floor, planning roof access maintenance or top-level evacuation procedures.
    \item \textbf{Doors and windows per storey}: A maintenance manager might query the IFC model for the number of doors and windows per storey to schedule inspections or replacements by floor.
    \item \textbf{Bottom floor slabs/ceilings}: A structural engineer may query the IFC model to identify the slab elements belonging to the lowest floor in order to review load transfer to the foundation system and verify the structural layout.
    \item \textbf{Storey with fewest rooms}: A facility manager might query the IFC model to identify the storey with the fewest rooms to determine whether it is primarily used for mechanical services, equipment spaces, or other non-occupiable functions.
    \item \textbf{Rooms and their volumes}: An HVAC engineer or space planner might retrieve each room’s volume from the IFC model to calculate heating and cooling loads or analyze usable space.
    \item \textbf{Rooms connected to room “X”}: A facility manager or safety and security planner might query the IFC to find which rooms are directly connected to room “X” to plan access control or service routes.
    \item \textbf{Room with most doors}: A fire safety officer might query the IFC to identify which room has the most doors in order to ensure it is properly marked and managed for evacuation routes.
    \item \textbf{Centroid of room “X”}: An indoor navigation or robotics system might query the IFC model for the centroid of room “X” to obtain a representative reference point within the room for positioning, navigation targets, or spatial calculations.
    \item \textbf{Distance between rooms “A” and “B”}: Emergency and rescue teams could query the IFC model for the distance between rooms during evacuation planning to estimate travel time and ensure rapid response.
    \item \textbf{Rooms connected by doors to “X lobby”}: An event organizer or facilities manager might query the IFC to list which rooms have doors leading to the “X lobby” to plan signage and guide visitors from the lobby to their destinations.
    \item \textbf{Bounding box of room “X”}: A BIM analyst might retrieve the bounding box of room “X” from the IFC model to quickly approximate its spatial extent when performing spatial indexing, collision checks, or coarse spatial analysis.
    \item \textbf{Total doors in building}: A facilities manager or maintenance planner might query the IFC model to find the total number of doors in the building for inventory management or replacement scheduling.
    \item \textbf{External vs. internal walls by floor}: An energy engineer might query the IFC model to identify which walls are external versus internal (organized by floor) to assess insulation requirements and thermal performance.
    \item \textbf{Beams/columns per storey}: A structural engineer or materials planner could query the IFC for the count of beams or columns on each storey to prepare structural calculations or order materials.
    \item \textbf{Available roofs}: An architect or maintenance team might query the IFC to list what roof types or roof elements are present (e.g. flat roof) to verify design options or maintenance plans.
    \item \textbf{Doors’ materials/fire ratings}: A fire safety inspector could retrieve from the IFC all doors’ materials and fire ratings to ensure they comply with fire safety regulations.
    \item \textbf{Navigable path from Lobby to room “X”}: A security planner or robotic guide system might query the IFC model to determine if a navigable path exists from the main lobby to room “X,” verifying accessibility or planning a route.
    \item \textbf{Shortest path between rooms “X” and “Y”}: A facility planner or indoor navigation system might query the IFC model to compute the shortest path between room “X” and room “Y” to guide occupants or service staff through the most efficient route inside the building.
    \item \textbf{Isolated rooms}: A building safety auditor might query the IFC model to identify rooms that are not reachable from other spaces to detect potential evacuation or accessibility issues in the building layout.
    \item \textbf{Fire rating of door “X”}: A fire safety engineer would query the IFC model for the fire rating of a specific door “X” to confirm it meets code requirements for its location.
    \item \textbf{Wall properties}: A BIM manager or architect might want to know the wall properties available in the IFC model to verify that IFC is properly created and  critical information (e.g. material, thickness, fire rating) is present for each wall. 
    \item \textbf{U-value of external walls}: An energy consultant or HVAC engineer could query the IFC model for the U-value (thermal transmittance) of the external walls to assess the building’s energy performance and ensure compliance.
    \item \textbf{EXIT doors}: A fire marshal or safety inspector might query the IFC to identify which doors are designated as EXIT doors to verify they are correctly labeled and meet emergency egress standards.
\end{itemize}
\newpage

\end{document}